\documentclass{article}

\PassOptionsToPackage{numbers, compress}{natbib}

\usepackage[pagebackref,breaklinks,colorlinks]{hyperref}

\usepackage[preprint]{neurips_2022}

\usepackage[utf8]{inputenc} 
\usepackage[T1]{fontenc}    
\usepackage{hyperref}       
\usepackage{url}            
\usepackage{booktabs}       
\usepackage{amsfonts}       
\usepackage{nicefrac}       
\usepackage{microtype}      
\usepackage{xcolor}         
\usepackage{bm}
\usepackage[rightcaption]{sidecap}
\usepackage{caption}

\usepackage{amsmath}
\usepackage{amssymb}
\usepackage{mathtools}
\usepackage{amsthm}
\usepackage{gensymb}

\usepackage{enumitem}

\newcommand{\wj}[1]{{\color{black}#1}}

\title{On the Strong Correlation Between Model Invariance and Generalization}

\author{Weijian Deng~ ~  Stephen Gould~~  Liang Zheng\\
	 Australian National University \\
   {\tt\small ~~~\{firstname.lastname\}@anu.edu.au}
}
    
\begin{document}

\maketitle

\begin{abstract}
Generalization and invariance are two essential properties of any machine learning model. Generalization captures a model's ability to classify unseen data while invariance measures consistency of model predictions on transformations of the data. Existing research suggests a positive relationship: a model generalizing well should be invariant to certain visual factors. Building on this qualitative implication we make two contributions. First, we introduce effective invariance (EI), a simple and reasonable measure of model invariance which does not rely on image labels. Given predictions on a test image and its transformed version, EI measures how well the predictions agree and with what level of confidence. Second, using invariance scores computed by EI, we perform large-scale quantitative correlation studies between generalization and invariance, focusing on rotation and grayscale transformations. From a model-centric view, we observe generalization and invariance of different models exhibit \emph{a strong linear relationship}, on both in-distribution and out-of-distribution datasets. From a dataset-centric view, we find a certain model's accuracy and invariance \textit{linearly correlated} on different test sets. Apart from these major findings, other minor but interesting insights are also discussed.
\end{abstract}

\section{Introduction}\label{sec:intro}
Generalization and invariance are two important model properties in machine learning. The former characterizes how well a model performs when encountering in-distribution or out-of-distribution (OOD) test data \cite{shai2010hdh,miller2021accuracy,taori2020measuring,hendrycks2021many,zhang2021understanding}. 
The latter assesses how consistent model predictions are on transformed test data \cite{azulay2018deep,engstrom2019exploring,kanbak2018geometric,zhang2019making,kayhan2020translation,zhu2021understanding}. 
Therefore, understanding how the two properties are related would benefit model decision analysis under dynamic environments. 

However, there lacks \textit{quantitative} and \textit{systematic} analysis of the relationship between generalization and invariance. In fact, most existing works are qualitative. For example, adding rotation invariance to the model improves its in-distribution (ID) classification accuracy \cite{zhou2017oriented,jaderberg2015spatial,delchevalerie2021achieving}; \wj{a shift-invariant model is robust to perturbation \cite{zhang2019making}}. Moreover, existing research is limited to a few in-distribution datasets and classifier architectures. As such, the relationship of interest remains unknown on many other scenarios, such as out-of-distribution and large-scale test data, and other types of models.

Before performing the quantitative study, it is necessary to first quantify generalization and invariance. \wj{For the former, the deep models we consider are well trained, so we simply use the accuracy on the test set, as in many previous works \cite{taori2020measuring,miller2021accuracy,recht2019imagenet}}. In comparison, quantifying invariance is not as straightforward. Some works use model accuracy drop when the test set undergoes transformations to indicate invariance ability \cite{schiff2021predicting,engstrom2019exploring,delchevalerie2021achieving}. While this strategy is useful for a single model, its effectiveness is limited when comparing invariance of multiple models. Others resort to consistency, \textit{i.e.}, models should have the same decision \cite{zhang2019making,azulay2018deep}, but this method neglects prediction \textit{confidence}, which we find critical for describing invariance (see discussions in Section \ref{sec:ei}). 

We make two contributions to the community. \textbf{First}, we propose a new method to measure model invariance, named effective invariance (EI), which considers both the consistency and confidence of predictions. Given a test image and its transformed counterpart, if the model predicts the same class and with high confidence, the EI value or invariance strength is high. Otherwise, if the model makes different class predictions or the confidence is low, the EI score will be low. We show this new measure solves invariance valuation in canonical cases where the commonly used metrics (\textit{e.g.}, Jensen-Shannon divergence) may fail. \textbf{Second}, we conduct a broad correlation study to quantitatively understand the relationship between model generalization and invariance. Specifically, we use $8$ test sets with various distribution types, such as the in-distribution ImageNet validation set \cite{deng2009imagenet}, and out-of-distribution ImageNet-Rendition with style shift \cite{hendrycks2021many}.  We evaluate $150$ ImageNet models ranging from traditional convolution neural networks (\emph{VGGs} \cite{simonyan2014very}) to the very recent vision transformers (\emph{e.g.}, BEiT \cite{bao2021beit}). Below we list two key observations and example insights.

\begin{itemize}[leftmargin=*]
\item For \emph{various models}, there is a strong correlation between their accuracy and invariance on both in-distribution and out-of-distribution datasets (Sections \ref{sec:rot} and \ref{sec:grayscale}). This finding can be useful for unsupervised model selection, because EI does not require test ground truths.
\item On \emph{various out-of-distribution datasets}, a model's accuracy and EI scores are also strongly correlated (Section \ref{sec:autoeval}). 
This observation can be used to predict model accuracy on out-of-distribution datasets without access to ground truths.
\item Compared with data augmentation, training with more data seems more effective to improve invariance and generalization (Section \ref{sec:augmention}). 
\end{itemize}

\section{Related Work}\label{related}
\textbf{Predicting generalization gap: a model-centric view.} This task aims to predict the generalization gap of machine learning models on in-distribution data, \emph{i.e.}, the difference between training and test accuracy.
Most existing works focus on developing \emph{complexity measurement} of trained network parameters and training data~\cite{eilertsen2020classifying,unterthiner2020predicting,arora2018stronger,corneanu2020computing,jiang2018predicting,neyshabur2017exploring,jiang2019fantastic,garg2021ratt,jiang2021assessing}, such as persistent topology \cite{corneanu2020computing} and the product of norms of the weights across layers \cite{neyshabur2017exploring}.
These methods, assuming the training and test distributions are the same, \emph{do not} consider the characteristics of the test distribution. 
Thus, they have limited effectiveness to predict generalization under out-of-distribution test sets.
In addressing this limitation, we show that invariance, under our definition, serves as a strong indicator of model generalization ability or accuracy on both in-distribution and out-of-distribution test data. 

\wj{Closely related to our work, two recent methods \cite{kashyap2021robustness,schiff2021predicting} predict ID generalization gap based on how a network performs on perturbed data points. Specifically, Kashyap \emph{et al.} \cite{kashyap2021robustness} 
use confidence drop to represent invariance, which is less effective for invariance measurement under OOD data. Schiff \emph{et al.}~\cite{schiff2021predicting} uses accuracy drop to measure invariance, which needs test labels, while EI does not require test labels and is more reasonable than accuracy under OOD data. Besides, drawing a response curve in \cite{schiff2021predicting} is computationally heavy, while our method is efficient. 
Further, both studies are limited in their scope: they mainly study ID generalization, has two types of networks and lack large-scale test sets, while our work is much more comprehensive.}

\textbf{Predicting generalization gap: a dataset-centric view.} The overall goal of this task is to predict the performance of a given model on various unlabeled test sets~\cite{deng2021labels,guillory2021predicting,garg2022leveraging,baek2022agreement,ng2022predicting,yu2022predicting}. Many methods take into account the statistics of the test set for accuracy prediction \cite{deng2021labels,guillory2021predicting,Deng:ICML2021,garg2022leveraging,chen2021detecting}, such as distribution shift \cite{deng2021labels}, average Softmax score on each test set \cite{guillory2021predicting}.
We contribute a new solution: using the model's invariance on the OOD dataset to predict its accuracy. This is supported by our new observation of strong linear correlation between a certain model's accuracy and invariance on various test sets.

\textbf{Improving robustness with data augmentation.} Data augmentation transforms training data to increase its diversity, which helps learn more robust models \cite{hendrycks2019augmix,yun2019cutmix,cubuk2018autoaugment,cubuk2020randaugment,devries2017improved,mintun2021interaction,hendrycks2021pixmix}.
For example, Mixup \cite{zhang2017mixup,tokozume2018between} and AutoAugment \cite{cubuk2018autoaugment} are shown to improve model performance under distribution changes \cite{yin2019fourier,mintun2021interaction}.

Instead of using common transformations, adversarial training \cite{madry2017towards,salman2020adversarially,rusak2020simple} augments training images with an adversarially learned noise distribution. 
While these works aim to improve corruption robustness with data augmentation, we instead use the latter to analyze model invariance (and generalization).

\section{Proposed Effective Invariance (EI)}\label{sec:ei}
\textbf{Notations}. Considering an $K$-way classification task, we define input space $\mathcal{X} \in \mathbb{R}^{d}$ and label space $\mathcal{Y}=\{1,..., K\}$. Given a sample $(\bm{x}, y)$ drawn from an unknown distribution $\pi$ on $\mathcal{X} \times \mathcal{Y}$, a neural network classifier $\bm{f}: \mathbb{R}^{d} \to \Delta_K$ produces a probability distribution for $\bm{x}$ on $K$ classes, where $\Delta_K$ denotes the $K-1$ dimensional unit simplex. Specifically, ${f}_{i}(\bm{x})$ denotes the $i$-th element of the Softmax output vector produced by $\bm{f}$. Then, $\hat{y} \eqqcolon \arg\max_{i}{f}_{i}(\bm{x})$ is the predicted class, and $\hat{p} \eqqcolon \max_i {f}_{i} (\bm{x})$ is the associated confidence score. 
Image transformation is defined as $\bm{\mathcal{T}}: \mathbb{R}^{d} \to \mathbb{R}^{d}$. Then, the transformed image is $\bm{x'} = \bm{\mathcal{T}}(\bm{x})$, and its predicted class is $\hat{y}_{t} \eqqcolon \arg\max_{i}{f}_{i}(\bm{x'})$ with confidence score $\hat{p}_{t} \eqqcolon \max_i {f}_{i} (\bm{x'})$.  

\begin{figure*}
    \centering
    \includegraphics[width=1.0\linewidth]{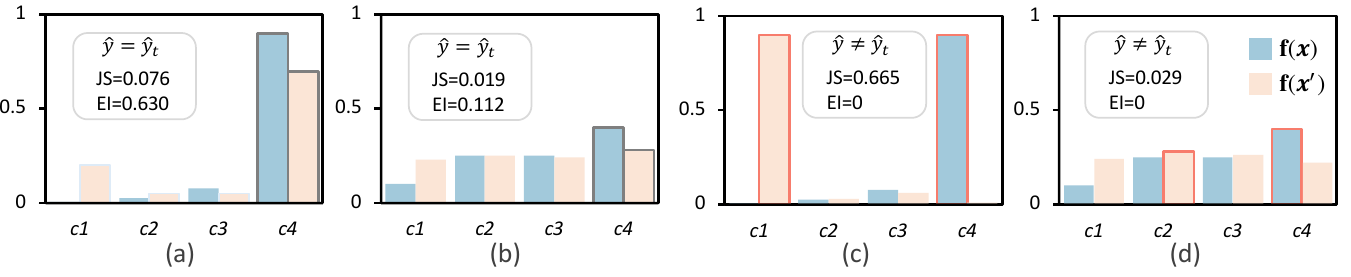}
    \caption{\textbf{An illustrative comparison of EI and Jensen-Shannon divergence (JS) as invariance measures.} Four representative cases are shown, where for each class ($c1$-$c4$) we show the Softmax output of the original image $\bm{f}(\bm{x})$ and the transformed image $\bm{f}(\bm{x'})$.
    On the one hand, the model exhibits higher invariance in \textbf{(a)} than \textbf{(b)}, because it makes the same class predictions ($\hat{y} = \hat{y}_{t}$) and has higher confidence in (a). This is correctly reflected by EI ($0.630$ \emph{vs.} $0.112$;  higher is better), but JS incorrectly decides the opposite way ($0.076$ \emph{vs.} $0.019$; lower is better), because JS does not consider confidence explicitly. 
      In cases \textbf{(c)} and \textbf{(d)}, the model makes different class predictions ($\hat{y} \neq \hat{y}_{t}$), so its invariance should be very low. This is again correctly captured by EI (0 value for both cases), but JS erroneously gives high invariance to (d), due to the fact that JS merely looks at the global shape of the Softmax vectors without explicitly considering class prediction consistency. 
    }\label{ei}
    \vspace{-0.4cm}
\end{figure*}

\textbf{Drawback of existing invariance measures}. A commonly seen strategy is to directly use the distance between the Softmax vectors of two predictions as invariance measure: a lower distance means higher invariance, and vice versa. Examples of the similarity metrics are 
Jensen-Shannon divergence (JS) \cite{fuglede2004jensen,hendrycks2019augmix} and $\ell_{2}$ distance \cite{sohn2020fixmatch,berthelot2019mixmatch} and Kullback–Leibler divergence \cite{kullback1951information,sun2021certified}. However, they only leverage the global similarity between two Softmax vectors without an explicitly consideration of prediction class consistency and confidence. 
We illustrate this drawback taking JS divergence as an example in Fig. \ref{ei}.
In cases (a) and (b) where the predicted classes are both consistent, JS decides classifier $\bm{f}$ in (b) has higher, which ignores the low confidence in (b).
In cases (d) where the predicted classes are different, JS still gives high invariance (small JS score), indicating a clear error.

\textbf{Definition of EI.} In its definition, given an image and its transformed sample, {a model with high invariance should give the same predicted class, and vice versa \cite{azulay2018deep,zhang2019making}}. In our definition of EI, we further use the prediction confidence. Our motivation is as follows. When a model predicts the same class for the two images, if either of the two predictions is of low confidence, we should not consider it as highly invariant but give a penalty. A model should have a high invariance if and only if it is highly confident in predicting the same class. 
Based on these considerations, EI is defined as: 
\begin{equation}
    EI =
    \begin{dcases}
        \sqrt{\hat{p}_{t}\cdot\hat{p}} & \text{if }\hat{y}_{t} = \hat{y} \,;\\
        0 & \text{otherwise} \,. \\
    \end{dcases}
\end{equation}\label{eq:ei}

To better understand the soundness of EI, we depict four representative cases in Fig. \ref{ei}. 
In cases (a) and (b), classifier $\bm{f}$ gives the same predicted class ($\hat{y} = \hat{y}_{t}$) on original and transformed images. 
Under EI, Classifier $\bm{f}$ has a higher invariance ability in (a) because it has high confidence scores. In case (c), the predicted class are different ($\hat{y} \neq \hat{y}_{t}$), and in (d), the predictions are low in confidence (and give different classes), so we define invariance in both cases to be 0. Note that while we use the geometric mean of $\hat{p}_{t}$ and $\hat{p}$, our preliminary experiment shows the arithmetic gives similar effect.

\textbf{Computation of EI in practice.} Given a test image, we generate a transformed image using a certain transformation. Then, we compute the EI score based their Softmax vectors (Eq. \ref{eq:ei}).
We obtain the model invariance by averaging the EI scores over all the test images.
In this work, we mainly investigate the rotation transformation and grayscale transformation.
For the former, to avoid interpolation that would introduce artifact, we only use three transformation angles ($90 \degree, 180 \degree, 270 \degree$). For each rotation angle, we compute an overall invariance score by comparing with the predictions of original data. By averaging the three EI scores, we obtain rotation invariance on the test set.
{For the grayscale transformation, we remove color information and keep only luminous intensity information. Then, we compare with the predictions of grayscale and original data and compute the overall grayscale invariance on each test set. 
}

\section{Experimental Setup}\label{sec:exp}

\subsection{Models to Be Evaluated} \label{sec:models}

We considers both very recent and classic image classification models with different architectures, including {Convolutional Neural Networks} (\emph{e.g.}, standard VGGs \cite{simonyan2014very}, ResNets \cite{he2016deep}, and modern ConvNeXt \cite{liu2022convnet}), {Vision Transformers} (\emph{e.g.}, ViTs \cite{dosovitskiy2020image}, Swin \cite{liu2021swin}, and BEiT \cite{bao2021beit}), and {all-MLP architectures} \cite{ding2021repmlp,tolstikhin2021mlp} (\emph{i.e}, MLP-Mixer \cite{tolstikhin2021mlp}).

In addition to different architectures, we also cover models with various {training and regularization strategies} (\emph{e.g.}, learning rate schedule \cite{goyal2017accurate}, label smooth \cite{szegedy2016rethinking} and data augmentation \cite{hendrycks2019augmix,yun2019cutmix,cubuk2020randaugment,cubuk2018autoaugment}), {scaling strategies in model dimension}  (width, depth and resolution) \cite{bello2021revisiting,tan2019efficientnet,tan2021efficientnetv2}, and {learning manners} (supervised learning, semi-supervised learning \cite{yalniz2019billion} and knowledge distillation \cite{hinton2015distilling,heo2021rethinking}).
In total, we have {\textbf{{150 models}}} provided by TIMM \cite{rw2019timm}. They are either trained or fine-tuned on the ImageNet-1k training set \cite{deng2009imagenet}. 
The selected models can be roughly divided into the following {three} categories:

{{\textbf{Standard neural networks.}}} This category includes $100$ models only trained on ImageNet training set. These networks cover various architectures ranging from VGGs \cite{simonyan2014very} to EfficientNet \cite{tan2019efficientnet}.

{{\textbf{Semi-supervised learning.}}} We include $15$ models trained in a semi-supervised learning manner. They leverage a large collection of unlabelled images of YFCC100M \cite{thomee2015new} or Instagram 900M \cite{mahajan2018exploring} to improve the performance. We use models trained based on a teacher-student paradigm (\emph{e.g.}, SWSL-ResNet \cite{yalniz2019billion} and SSL-ResNet  \cite{yalniz2019billion}). 
Models trained with self-training methods on unlabeled JFT-300M \cite{sun2017revisiting} (\emph{e.g}, EfficientNet-L2-NS \cite{xie2020self}) are also included.

\textbf{Pretraining on more data.} We use another $35$ models that are pre-trained on significantly larger datasets than the standard ImageNet training set. 
Specifically, we consider three pre-training methods:
a) weakly supervised pretraining on \emph{IG-3.6B} (\emph{i.e.}, RegNetY \cite{singh2022revisiting} and ResNeXt101-WSL \cite{mahajan2018exploring}); b) supervised  pre-training on ImageNet-21K \cite{deng2009imagenet} (\emph{e.g.}, BiT \cite{kolesnikov2020big} and Swin \cite{liu2021swin}); (c) supervised pretraining on JFT-300M \cite{sun2017revisiting} (\emph{e.g.}, ViT L/16 \cite{dosovitskiy2020image}).

\subsection{Test Sets}
We use both {in-distribution (ID)} and {out-of-distribution (OOD)} datasets for the correlation study.
Specifically, the ImageNet validation set (ImageNet-Val) is used as {{ ID test set}}.
For {{OOD test sets}}, we use of {seven} datasets, each with a different distribution from standard ImageNet. Their distribution shift can be divided into the following {five} types.

{{\textbf{Dataset reproduction shift.}}} ImageNet-V2 \cite{recht2019imagenet} is a recollected version of ImageNet-Val.
It contains three versions resulting from different data sampling strategies : Matched-Frequency (A), Threshold-0.7 (B), Top-Images (C). Each version has $10,000$ images from $1,000$ classes.
\\
{{\textbf{Natural adversarial shift.}}} ImageNet-Adv(ersarial) \cite{hendrycks2021natural} is adversarially selected to be misclassified by ResNet-50. Its natural adversarial examples are unmodified real-world images and have been shown to be hard for other models \cite{hendrycks2021natural,taori2020measuring}. It has $7,500$ samples from $200$ ImageNet classes. 
\\
\textbf{Sketch shift.} ImageNet-S(ketch) \cite{wang2019learning} consists of sketch-like images and  matches ImageNet-Val in categories and scale. It contains $50,000$ images and shares the same $1,000$ classes as ImageNet.
\\
\textbf{Blur shift.} We use ImageNet-Blur with the highest severity provided by \cite{hendrycks2019robustness}. This dataset is synthesized by blurring ImageNet-Val images with a Gaussian function.
\\
{{\textbf{Style shift.}}} ImageNet-R(endition) \cite{hendrycks2021many} contains various abstract visual renditions (\emph{e.g.}, art, paintings, and video game) of ImageNet classes. ImageNet-R has $30,000$ images of $200$ ImageNet classes.

\subsection{Correlation Measures} 
We use Pearson Correlation coefficient ($r$) \cite{benesty2009pearson} and Spearman's Rank Correlation coefficient ($\rho$) \cite{kendall1948rank} to measure the linearity and monotonicity between invariance and generalization, respectively. Both coefficients range from $[-1, 1]$. A value closer to $-1$ or $1$ indicates strong negative or positive correlation, respectively, and $0$ implies no correlation \cite{benesty2009pearson}.
To precisely show the correlation, we use logit axis scaling that maps the accuracy range from $[0,1]$ to $[-\infty, +\infty]$, following \cite{taori2020measuring, miller2021accuracy}. Unless noted otherwise, the correlation coefficients are calculated using invariance score and non-linearly scaled accuracy numbers.

\begin{figure*}
    \centering
    \includegraphics[width=1.0\linewidth]{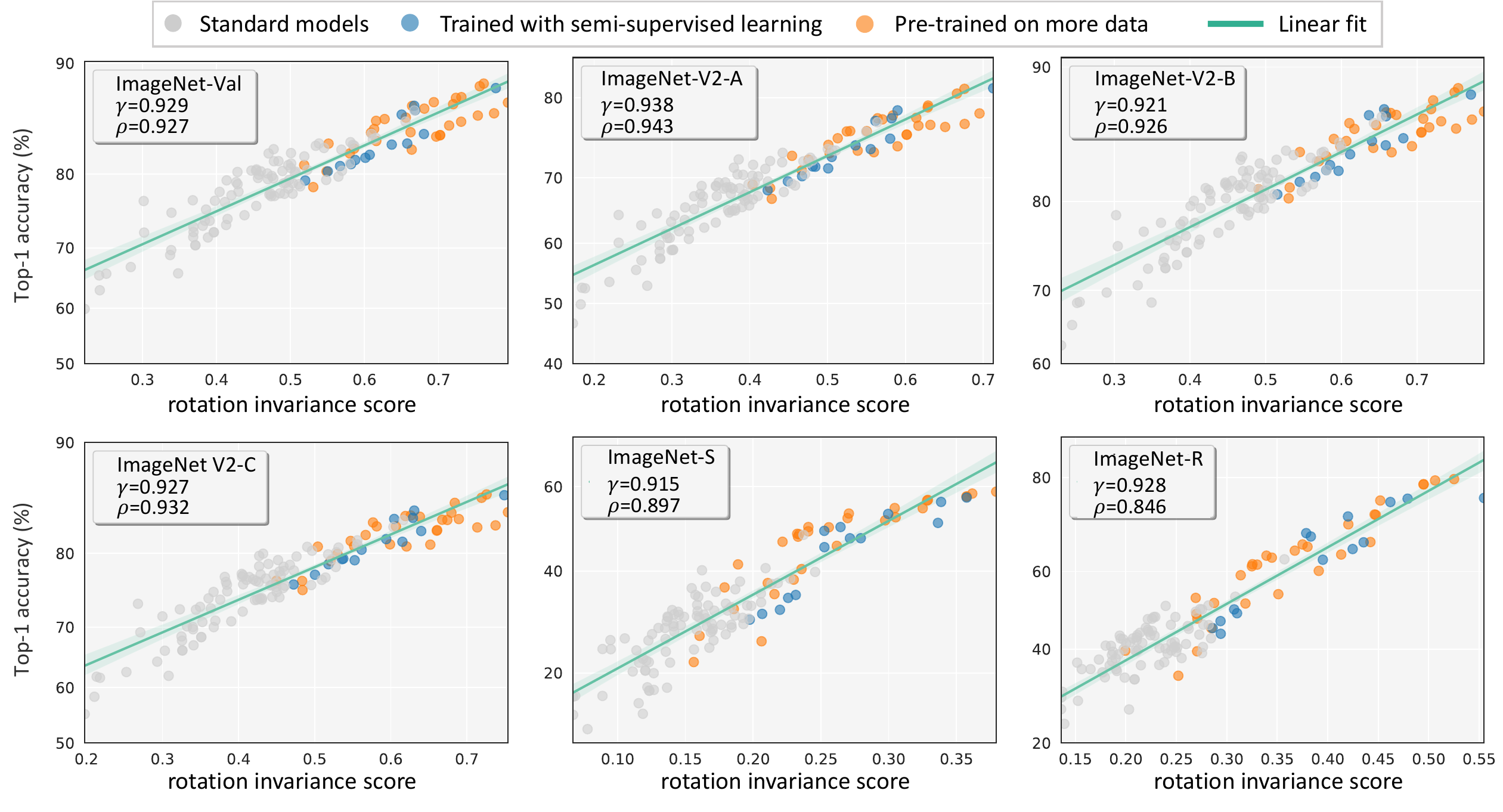}
    \caption{\textbf{Correlation between accuracy (\%) and rotation invariance (EI) for $150$ models.} Each figure is obtained from testing on a different ImageNet test set. In each figure, each dot denotes a model, and straight lines are fit by robust linear fit \cite{huber2011robust}. The shaded region in each figure is a $95\%$ confidence region for the linear fit from $1,000$ bootstrap samples. We clearly observe a strong linear relationship (Pearson's Correlation $r>0.915$ and Spearman's Correlation $\rho>0.875$). 
    }\label{rot}
\end{figure*}

\section{Experimental Observations}
Most of our experiment is from a model-centric perspective, where we investigate how different models' accuracy and invariance correlate, and series of observations are made (Section \ref{sec:rot} - Section \ref{sec:augmention}). We also adopt a dataset-centric view, where we study how a given model's generalization and invariance correlate on different OOD test sets (Section \ref{sec:autoeval}).

\subsection{Strong Correlation Between Model's Rotation Invariance and Accuracy} \label{sec:rot}
In Fig. \ref{rot}, we show the correlation results of rotation invariance and generalization. We have two observations. \textbf{First}, we find that {for different models, their rotation EI scores have a linear relationship with their classification accuracy.} The correlation holds for both ID test and OOD test sets, various architectures and training strategies. Specifically, both correlation metrics $\lambda$ and $\rho$ are higher than 0.840. This indicates models with higher accuracy numbers are most likely to have stronger rotation invariance (measured by EI), and vice versa. To our knowledge, it is a very early observation of the quantitative relationship between generalization and invariance (to a certain factor).

\textbf{Second}, {{training with more data benefits rotation invariance and generalization.}} Large datasets contain images with various geometric variations. When (pre)trained with large datasets, models ({blue and orange dots in Fig. \ref{rot}}) adapt to the rotation variations and gains stronger invariance, which, according to our study, likely means a higher generalization accuracy on ID and OOD test sets.

\subsection{Strong Correlation Between Model's Grayscale Invariance and Accuracy} \label{sec:grayscale}
\begin{figure*}
    \centering
    \includegraphics[width=1.0\linewidth]{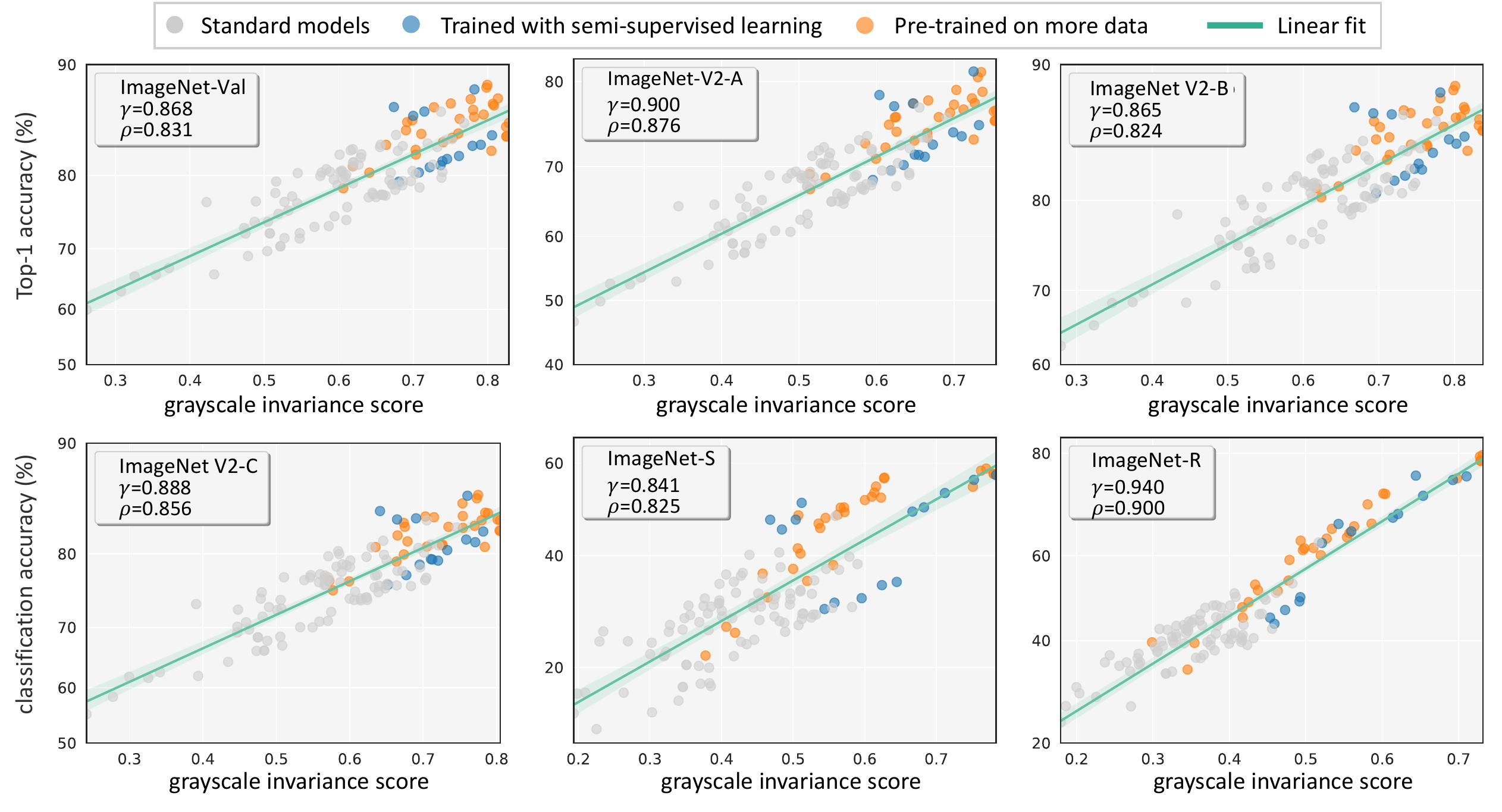}
    \caption{\textbf{Correlation between accuracy (\%) and grayscale invariance (EI) of $150$ models.} Similar to Fig. \ref{rot}, each dot denotes a model, where different colors denote different training strategies (see Section \ref{sec:models}). The subfigures correspond to three ImageNet test sets, respectively. 
    We observe the correlation is also strong: Pearson's Correlation $r>0.840$, and Spearman's Correlation $\rho>0.820$.
    }\label{grayscale}
    \vskip -0.1in
\end{figure*}

We now focus on grayscale invariance and report the correlation results in Fig. \ref{grayscale}. We have the following conclusions. \textbf{First}, among the $150$ models, there is strong linear correlation between accuracy and grayscale invariance measured by EI. Specifically, both correlation coefficients $r$ and $\rho$ are higher than $0.820$ on all test sets.
\textbf{Second}, we find that pretrained or semi-supervised models tend to have higher grayscale invariance and accuracy than standard models, which again indicates the usefulness of large training sets. 
\textbf{Third} and interestingly, the correlation is stronger on ImageNet-R than ImageNet-Val and ImageNet-V2-A ($0.940$ \emph{vs.} $0.900$ \emph{vs.} $0.868$). In fact, ImageNet-R is featured by style shift during its collection \cite{hendrycks2021many}, so for this test set being invariant to color changes is an important property for a stronger generalization ability.

\begin{figure*}
    \begin{center}
    \includegraphics[width=1\linewidth]{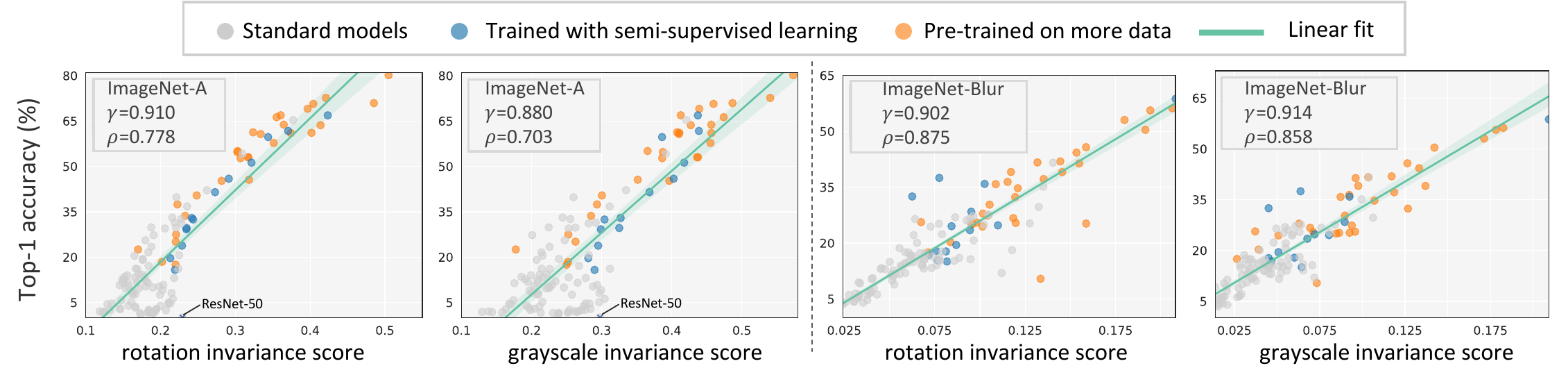}
    \caption{\textbf{Correlation study on very hard test sets.} We use the very hard ImageNet-A and Image-Blur for testing, where
    the accuracy of $83$ models is lower than $20\%$. Both rotation invariance and grayscale invariance are evaluated. 
    Overall we observe relatively solid correlation in all the four cases. Looking more closely, 
    most standard models (gray dots) are scattered in the low-accuracy region, while models trained with more data (blue and orange ones) move away from this region and exhibit linear trends. Thus, the overall linear correlation is high ($r >0.880$), and the overall rank correlation is slightly less consistent ($\rho$ ranges from $0.703$ to $0.858$) but still has clear trends. 
    }\label{hard}
    \end{center}
\end{figure*}

\subsection{{Correlation Exists on Very Hard Test Sets}} \label{sec:hardset}
We now study the correlation under very hard test sets and use ImageNet-A and ImageNet-GaussBlur for testing, on which accuracy of $83$ models is lower than $20\%$. We study rotation invariance and grayscale invariance. In Fig. \ref{hard}, we find strong linear correlation in the four cases ($r\leq0.88$). The rank correlation is less consistent but still indicates clear trends.
Moreover, most standard models have low accuracy, while models (pre)trained with more data tend to have high accuracy and invariance. 
We also notice standard models are scattered differently in the low-accuracy regime of ImageNet-A and ImageNet-Blur, possibly due to their dataset bias introduced during dataset construction \cite{meding2021trivial,hacohen2020let,yang2020towards}.

\begin{figure*}
    \centering
    \includegraphics[width=1.0\linewidth]{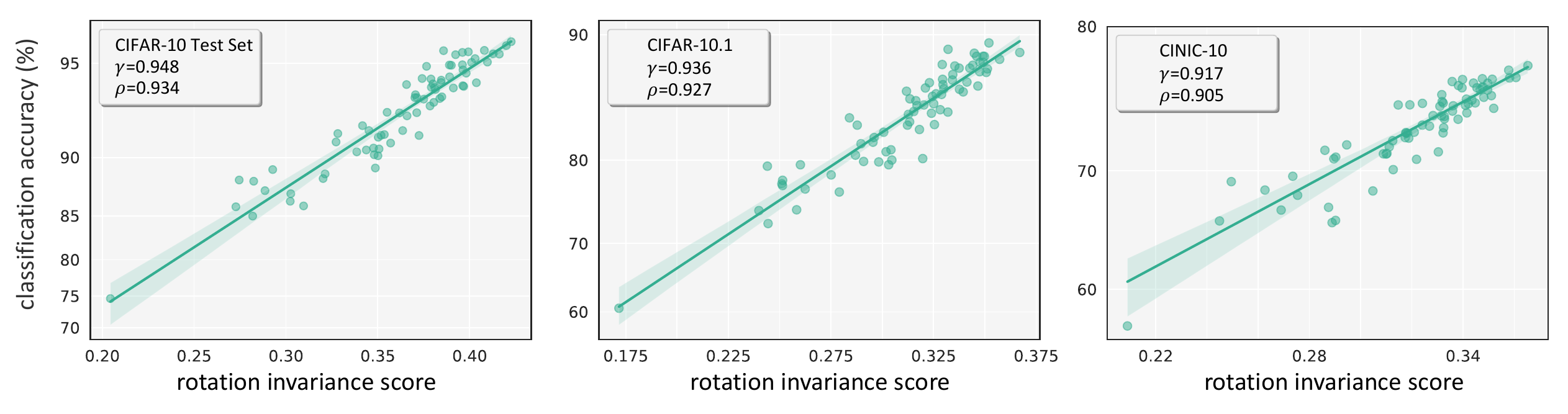}
    \caption{\textbf{Correlation study under the CIFAR-10 setup.} Each data point is a CIFAR model. Strong correlation exists between accuracy and rotation invariance on all the three test sets.
    }
    \label{cifar}
    \vskip -0.15in
\end{figure*}


\subsection{Correlation Holds Under the CIFAR-10 Setup} \label{sec:cifar}
We conduct correlation study on CIFAR-10 setup. We collect $90$ CIFAR models ranging from LeNet to EfficientNet. We use the ID {{CIFAR-10 test set}} and  two OOD test sets. 1) {{CIFAR-10.1}} \cite{recht2018cifar} is a reproduction of the CIFAR-10 test set but with distribution shift arising from changes in data collection. It contains $2,000$ test images sampled from TinyImages \cite{torralba200880}.
 2) {{CINIC-10 test set}} \cite{chrabaszcz2017downsampled} is an extended alternative for CIFAR-10. 
It has has $90,000$ images sampled from ImageNet. 

We show the relationship between model generalization and invariance in Fig. \ref{cifar}. On all the three test sets, we observe the strong correlation between model accuracy numbers and rotation invariance scores, where both $r$ and $\rho$ are greater than $0.90$.
{In the supplementary materials, we show correlation maintains with EI measured grayscale invariance.}

\subsection{EI Gives Stronger Correlation Than JS} \label{sec:EIvsJS}
In the representative cases (Fig. \ref{ei}), EI shows superiority to JS in measuring model invariance by explicitly considering Softmax consistency and confidence. Now we compare the invariance scores measured by EI and JS \emph{w.r.t} their correlation strength with accuracy in Fig. \ref{js-1} and Table \ref{js-2}. We find that JS provides good correlation on the ID ImageNet-Val test set, but much weaker correlation on the more difficult OOD tests than EI. As discussed in Section \ref{sec:ei}, the advantage of EI is that it not only follows the definition of invariance, but also integrates confidence to strength it, while JS only compares the overall Softmax vector. In fact, the drawback of JS is primarily reflected in hard test sets, where the Softmax vector usually has a flat shape. This explains why JS does not give strong correlation on the OOD test sets (see Section \ref{sec:hardset} for EI's performance on very hard test sets). We refer readers to the supplementary material for comparisons with other measures.

\begin{table}[t]
    \begin{minipage}{0.55\linewidth}
    \includegraphics[width=1.0\linewidth]{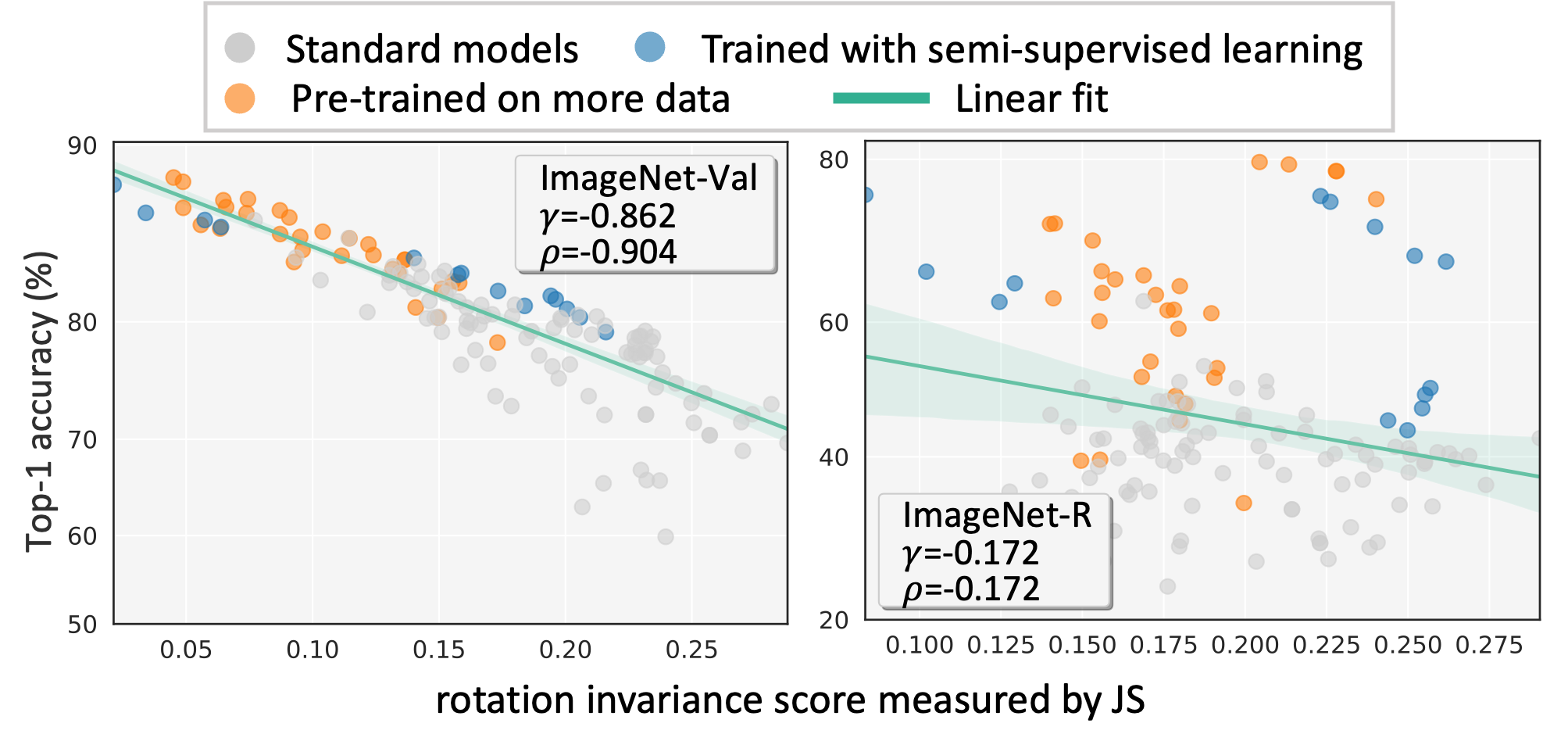}
     \captionof{figure}{\textbf{Correlation between accuracy (\%) and invariance (JS)}. Despite of the good correlation on ImageNet-Val, correlation on ImageNet-R, a harder test set, is weak. More results and analysis are provided in the supplementary meterials.}
     \label{js-1}
    \end{minipage}
    \quad
    \begin{minipage}{0.4\linewidth}
    \begin{small}
    \setlength{\tabcolsep}{8.5pt}
    \begin{tabular}{l|c|c}
    \toprule    
    Test set &  JS & EI \\ \midrule
    ImageNet-Val  & $-0.861$ & $+0.929$ \\ \midrule
    ImageNet-R    & $-0.173$ &  $+0.928$  \\ \midrule
    ImageNet-S    & $+0.131$ & $+0.915$    \\ \midrule
    ImageNet-A    & $-0.297$ & $+0.910$     \\ 
    \bottomrule 
    \end{tabular}
   \end{small}
   \vspace{0.4cm}
   \caption{\textbf{Comparing EI and JS.} 
   Under Pearson's Correlation $r$, JS does not show strong correlation on most datasets, while EI does on all the four datasets.}
   \label{js-2}
    \end{minipage}
    \vskip -0.2in
\end{table}

\begin{figure*}
    \centering
    \includegraphics[width=1.0\linewidth]{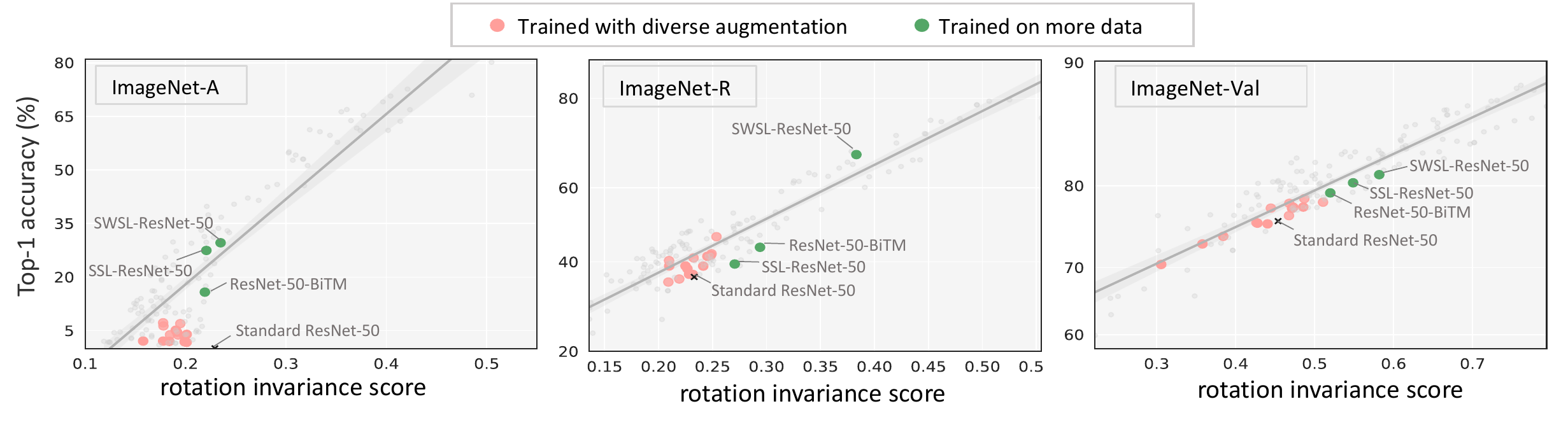}
    \caption{\textbf{Comparing data augmentation and (pre)training with more data.} We use $14$ ResNet-50 models (red dots) trained with various types of augmentation methods, and $3$ ResNet-50 models (green dots) trained on more data (real-world). 
    We also mark ResNet-50 trained with the standard learning strategies.
    On ImageNet-Val and ImageNet-R, we observe models trained with diverse augmentation follow a linear trend. However, they deviate from the trend on ImageNet-A. 
    In comparison, training with more data allows models to achieve relatively high accuracy and invariance on three test sets. 
    }\label{augmentation}
    \vskip -0.15in
\end{figure*}

\subsection{Comparing Two Training Manners of Their Generalization and Invariance} \label{sec:augmention}

Existing works report that using more diverse training data artificially (\emph{i.e.}, data augmentation) or naturally (\emph{i.e.}, more real-world data) improves model invariance \cite{hendrycks2021many,hendrycks2019augmix,yun2019cutmix,cubuk2018autoaugment,cubuk2020randaugment,devries2017improved,mintun2021interaction,von2021self}. 
In this study we compare the two strategies of their generalization and invariance abilities. We use $14$ ResNet-50 models trained with strong data augmentation such as PixMix \cite{hendrycks2021pixmix} and AutoAugment \cite{cubuk2018autoaugment}. 
For comparison, we employ another $3$ ResNet-50 models that learn invariance using more training samples: SWSL-ResNet-50, SSL-ResNet-50, and ResNet-50-BiTM. 
In Fig. \ref{augmentation}, we observe that models with heavy and strong augmentation exhibit linear trends on ImageNet-Val and ImageNet-R, but deviate on ImageNet-A. 
In comparison, models (pre)trained with more data follow the linear trend on three test sets. The latter seem more effective in improving invariance and accuracy compared with data augmentation, especially on ImageNet-A. To thoroughly understand and extend this initial observation, we will conduct more comprehensive experiment on this specific point in future. 

\begin{figure*}
    \centering
    \includegraphics[width=1.0\linewidth]{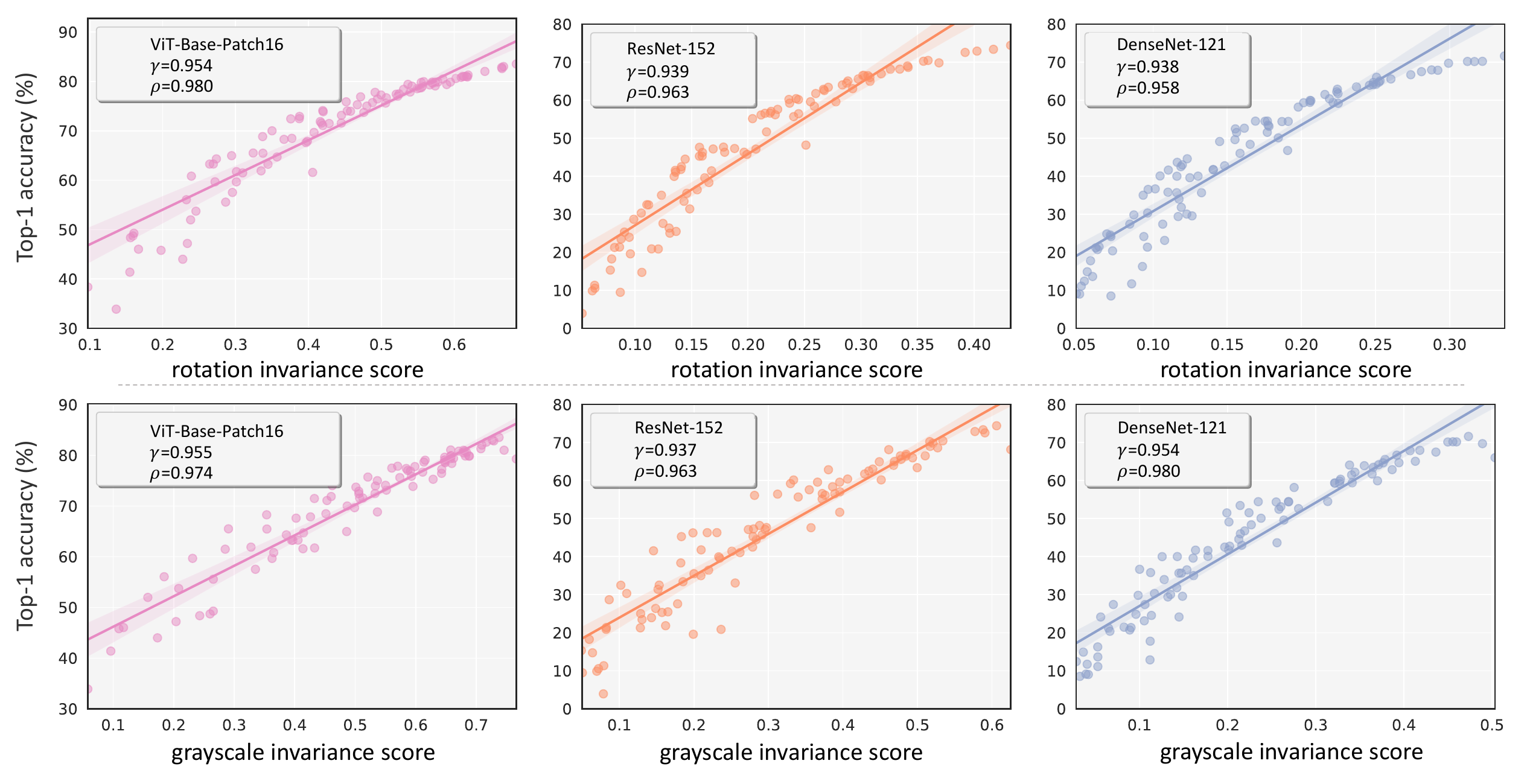}
    \caption{\textbf{{Correlation between a model's invariance and accuracy on various OOD test sets.}}
    In each figure, a data point corresponds to a test set from ImageNet-C \cite{hendrycks2019robustness}. The straight lines are fit with robust linear regression \cite{huber2011robust}.
    We test rotation invariance (top) and grayscale invariance (bottom). In each row, we use ViT-Base-Patch16, ResNet-152, and DenseNet-121, respectively. 
    In all sugfigures, we observe a strong negative correlation (Pearson's Correlation $r$ and Spearman's Rank Correlation $\rho$ are greater than $0.930$) between invariance and accuracy. 
    }\label{auto}
    \vskip -0.15in
\end{figure*}

\subsection{Correlation of A Model's Invariance and Generalization on Various OOD Test Sets}
\label{sec:autoeval}
From a \textbf{dataset-centric} perspective, we study the relationship between generalization and invariance of a given model on different OOD test sets. 
Specifically, given a model, we calculate its accuracy on every test test of ImageNet-C \cite{hendrycks2019robustness} and compute its rotation and grayscale invariance scores. 
We evaluate ResNet-152, ViT-Base-16, and DenseNet-121 classifiers for the correlation study.
As shown in Fig. \ref{auto}, there is a very strong correlation between classifier accuracy and invariance on various datasets ($r>0.93$ and $\rho>0.95$). The results indicate that a classifier tends to have a high accuracy on the test set where it has a high EI score. 
The above analysis indicates that it is feasible to use EI to access out-of-distribution error of a model.

\section{Conclusion}\label{sec:conclusion}
This work considers two critical properties of a machine learning model: invariance and generalization. To study their relationship, we first introduce effective invariance (EI) to more reasonably measure invariance and then provide in-depth and comprehensive correlation experiment.
From a model-centric perspective, we report accuracy and EI of various models has a strong linear relationship on both ID and OOD datasets, which is validated on many scenarios such as large-scale test sets, CIFAR-10 and very hard test sets. From a dataset-centric perspective, we show the accuracy and EI of a model have a strong linear relationship on various OOD datasets.

\textbf{Limitations and potential directions.} \textbf{First}, some networks with specially designed modules can be highly invariant to some transformations \cite{engstrom2019exploring,kanbak2018geometric,zhang2019making,kayhan2020translation,jaderberg2015spatial,weiler2019general}, such as rotation invariance networks \cite{zhou2017oriented,cohen2018spherical,delchevalerie2021achieving}.
The rotation invariance scores of these models may present very different correlation from our observations. That said, their color invariance scores may still exhibit similar correlation with this work. 
In future works, it would be interesting to understand how these models deviate from others.
\textbf{Second}, our ImageNet test sets do not include the \emph{geographic shift} where images are captured from various locations \cite{hendrycks2021many}. Recent works show this shift is also a key factor influencing model accuracy \cite{wilds2021}. We leave this question to future study. \textbf{Lastly}, this work focuses on classification tasks, where models are supervised by image-level annotations. In other computer vision tasks, models may be trained with different levels of supervision, such as instance-level bounding box annotations \cite{ren2015faster}, pixel-level labels \cite{long2015fully} and temporal context supervision \cite{wang2018non}. Different types of supervision may lead to invariance ability to different factors, which may exhibit different correlation profiles.

\textbf{Potential negative social impact.} We study fundamental model properties with public classification datasets, which might be misused in certain applications with ethical concerns.

\appendix

In the supplementary document, we first compare effective invariance (EI) with other metrics \emph{w.r.t} their correlation strength with accuracy in Section \ref{sec:measures}. 
Then, we show more correlation results under CIFAR-10 setup in Section \ref{sec:cifar-grayscale}. Moreover, we show more correlation results between a model’s invariance and generalization on out-of-distribution (OOD) datasets in Section \ref{sec:autoeval-more}.
Last, we include the details of models and datasets in Section \ref{sec:models}.

\section{Comparison With Other Measures}\label{sec:measures}
In this section, we compare the invariance scores measured by EI and other metric \emph{w.r.t} their correlation strength with accuracy. In the main paper, we compare EI with JS. Here, we additionally use the following five metrics.

(a) \textbf{$\ell_{2}$ distance} \cite{berthelot2019mixmatch,sohn2020fixmatch}. It defines the invariance as the $\ell_{2}$ distance between predictions of original and translated data. A smaller $\ell_{2}$ distance indicates a higher invariance score.

(b)\textbf{ Accuracy difference} \cite{schiff2021predicting}. The invariance is defined as the overall accuracy change on a test set after using image transformation on all test samples. A smaller difference indicates a higher invariance score.

(c) \textbf{Confidence only}. A model's average prediction confidence on all test samples is used as its invariance score. A higher confidence indicates a higher invariance score.

(d) \textbf{Prediction consistency only} \cite{azulay2018deep,zhang2019making}. It defines the invariance by only checking whether the predicted class is the same without considering the prediction confidence. For each test sample, if the predicted class is the same, the score is $1$; if the predicted classes are different, the score is $0$.

\begin{figure*}[!th]
    \centering
    \includegraphics[width=1.0\linewidth]{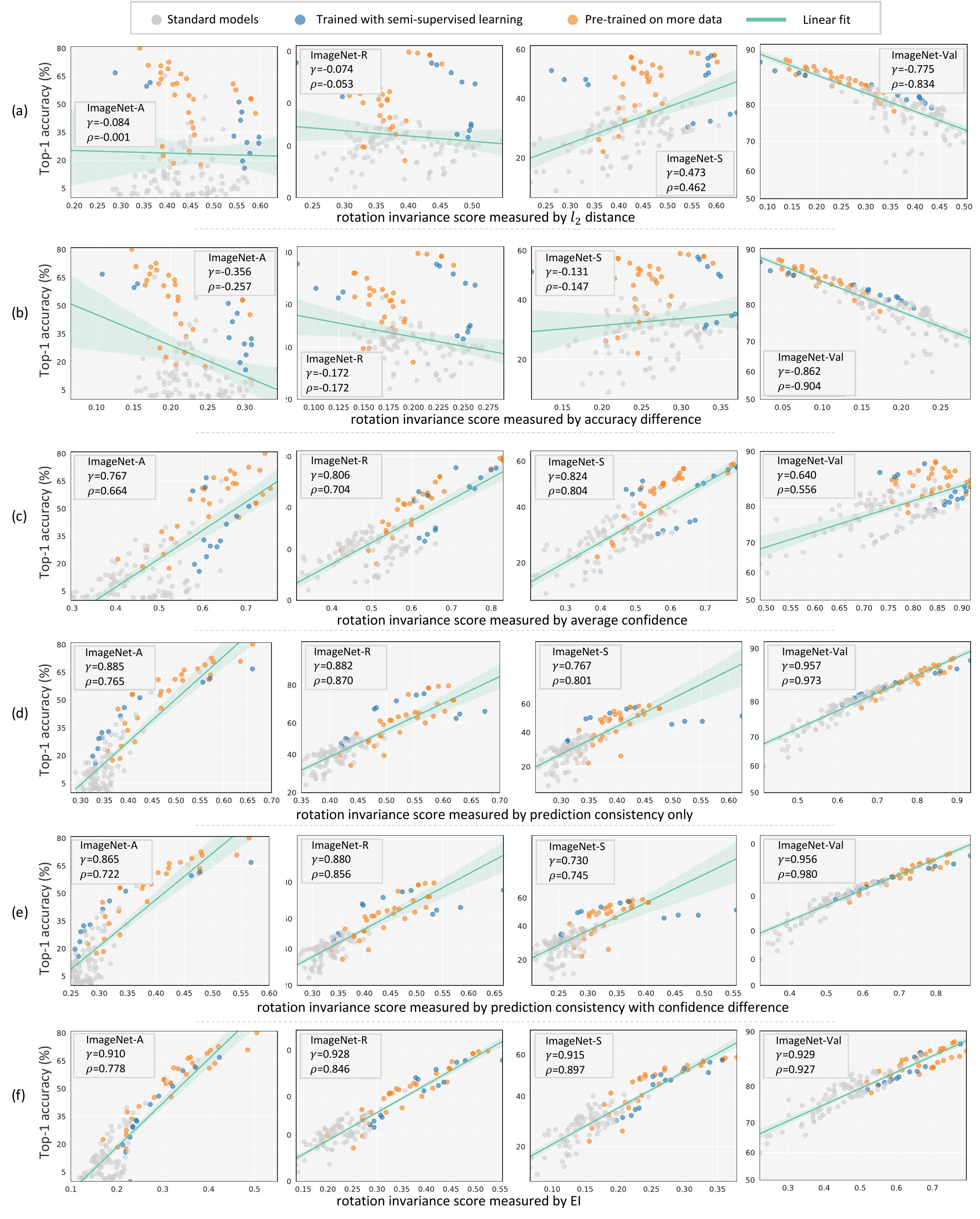}
    \caption{\textbf{Correlation between accuracy (\%) and rotation invariance.} We show the correlation results obtained using six invariance measures. From top to bottom: (a) $\ell_{2}$ distance, (b) accuracy difference, (c) confidence only, (d) prediction consistency only, (e) prediction consistency with confidence difference, and (f) effective invariance (EI). 
    Compared with other measures, the invariance score measured by EI provides a reasonably good correlation with model accuracy on ID ImageNet-Val and OOD datasets (ImageNet-S/A/R).
    }
    \vskip -0.15in
    \label{measures}
\end{figure*}

(e) \textbf{Prediction consistency with confidence difference} \cite{kashyap2021robustness}. It can be viewed as an variant of EI score. For each test sample, if the predicted class is the same, then the score is defined as $1-\|\hat{p}_{t}-\hat{p}\|$, where $\hat{p}$ and $\hat{p}_{t}$ is the prediction confidence on original data and transformed data, respectively; if the predicted classes are different, the score is $0$.

In Fig. \ref{measures}, we show the correlation results using different invariance measures. 
We find all measures give a clear correlation trendy on in-distribution (ID) ImageNet-Val. However, when testing on OOD datasets, $\ell_{2}$ distance and accuracy difference do not show good correlation trendy.
Moreover, using model's average confidence as the invariance score cannot show high correlation strength with accuracy. Compared with measuring invariance by prediction consistency only, EI provides relatively stronger correlation with model accuracy on OOD datasets. 
The above analysis shows that EI is superiority to other methods in measuring model invariance by showing relatively high correlation strength with model accuracy on both ID and OOD datasets.

\section{More Correlation Results Under CIFAR-10 Setup}\label{sec:cifar-grayscale}
\begin{figure*}[!th]
    \centering
    \includegraphics[width=1.0\linewidth]{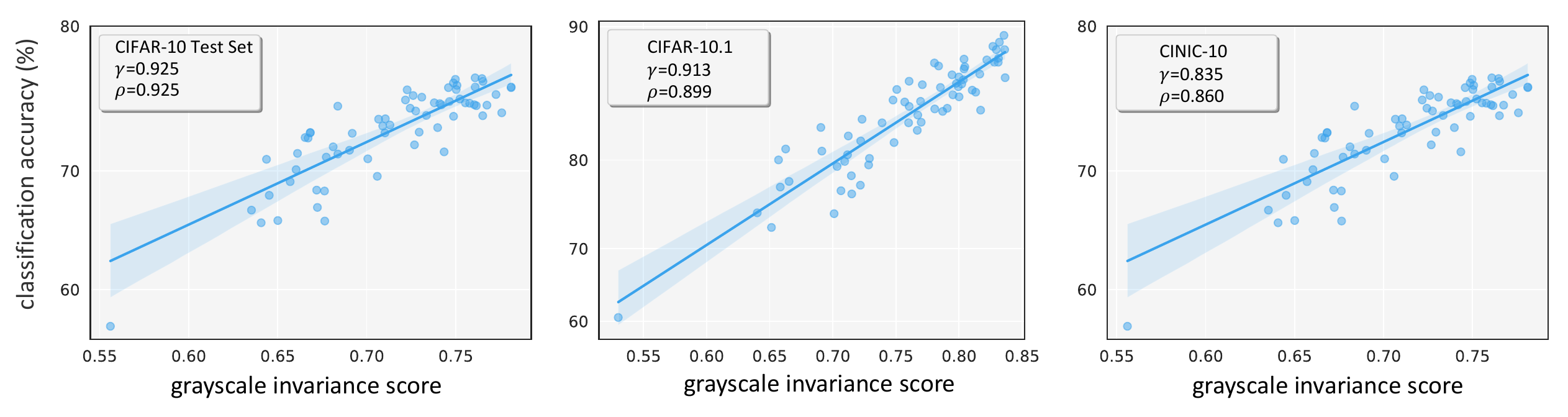}
    \caption{\textbf{Correlation study under the CIFAR-10 setup.} Each data point is a CIFAR model. We show that strong correlation exists between accuracy and grayscale invariance on all the three datasets (CIFAR-10 test set \cite{krizhevsky2009learning}, CIFAR-10.1 \cite{recht2018cifar}, and CINIC-10 test set \cite{chrabaszcz2017downsampled}). 
    The y-axis is adjusted using logit scaling and the robust linear fit \cite{huber2011robust} is computed in the scaled space.
    The shaded region in each panel is a 95\% confidence region for the linear fit from $1,000$ bootstrap samples.
    }
    \label{appendix:cifar}
\end{figure*}

In Fig. \ref{appendix:cifar}, we additionally show the correlation between grayscale invariance and accuracy. We observe that the correlation remains under CIFAR-10 setup: the correlation is relatively strong on each test set: Peasrson's Correlation ($r$) and Spearman's Rank Correlation ($\rho$) are larger than $0.830$.

\section{More Correlation Results Between a Model’s Invariance and Generalization on OOD Datasets} \label{sec:autoeval-more}

\begin{figure*}[!th]
    \centering
    \includegraphics[width=1.0\linewidth]{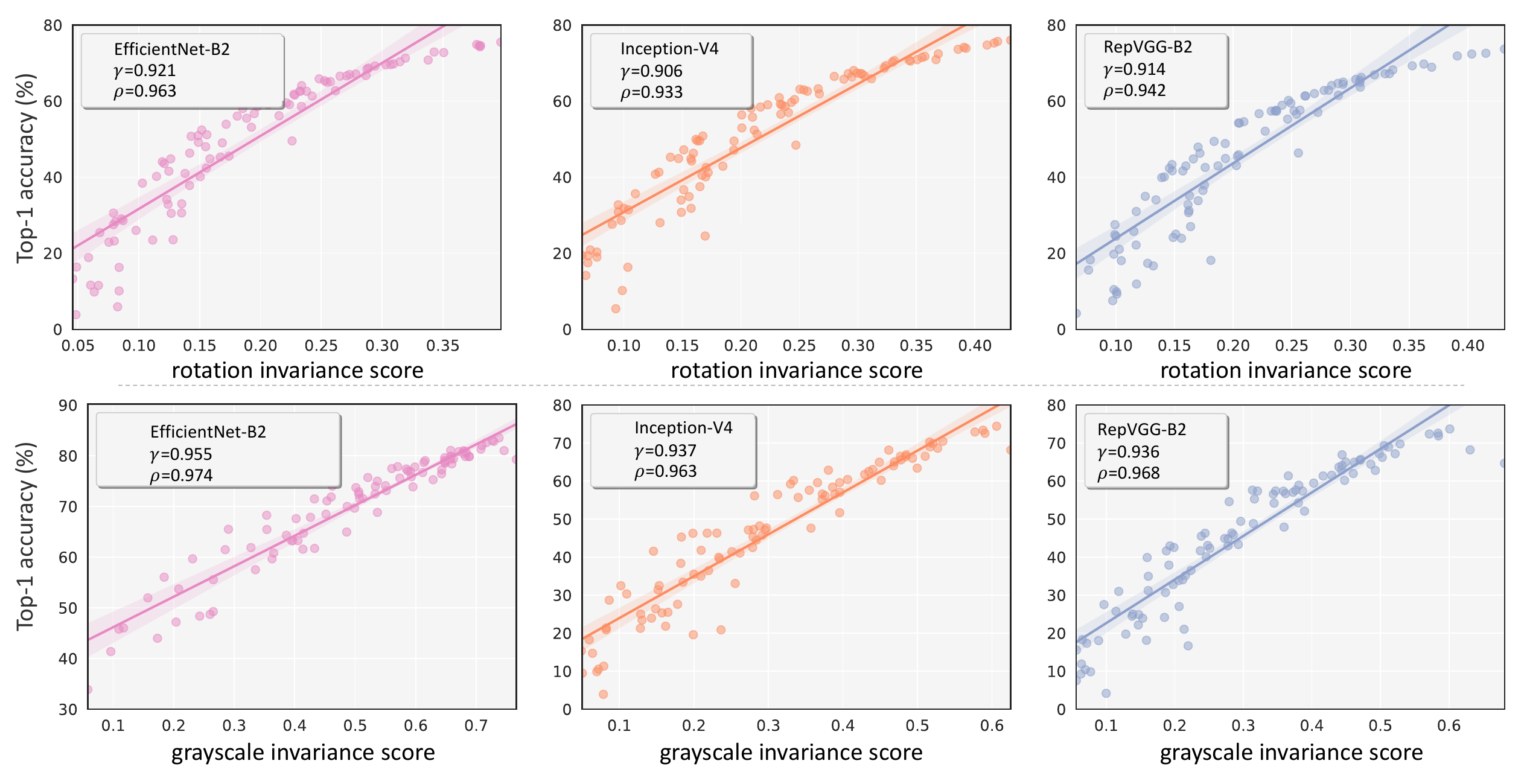}
    \caption{\textbf{{Correlation between a model's invariance and accuracy on various OOD test sets.}}
    In each figure, a data point corresponds to a test set from ImageNet-C \cite{hendrycks2019robustness}. The straight lines are fit with robust linear regression \cite{huber2011robust}.
    We test rotation invariance (top) and grayscale invariance (bottom). In each row, we use EfficientNet-B2, Inception-V4, and RepVGG-B2, respectively. 
    In all sugfigures, we observe a strong negative correlation (Pearson's Correlation $r$ and Spearman's Rank Correlation $\rho$ are greater than $0.900$) between EI invariance score and model accuracy. 
    The shaded region in each panel is a 95\% confidence region for the linear fit from $1,000$ bootstrap samples.
    }
    \label{supp-autoeval}
\end{figure*}

In this section, we additional report the correlation results using other three classifiers: EfficientNet-B2, Inception-V4, and RepVGG-B2. 
As shown in \ref{supp-autoeval}, there is a very strong correlation between classifier accuracy and invariance score measured by EI on various datasets ($r > 0.900$ and $\rho > 0.930$). 
This indicate that a classifier tends to have a high accuracy on the test set where it has a high EI score. 
The above observation further shows that it is feasible to use EI score to access out-of-distribution error of a model without using labels.

\section{Experimental Setup}

\subsection{ImageNet Models} 
In total, we use ImageNet models provided by PyTorch Image Models (timm) \cite{rw2019timm}. They are either trained or fine-tuned on the ImageNet-1k training set \cite{deng2009imagenet}. 
The models are listed in the following.

\textbf{(1) Standard neural networks}
\\
\emph{\{
`resmlp\_36\_224',
`cait\_s36\_384',
`cait\_s24\_224',
`convit\_base',
`convit\_tiny',
`twins\_pcpvt\_base',
`eca\_nfnet\_l1',
`xcit\_tiny\_24\_p8\_384\_dist', 
`efficientnet\_b1',
`efficientnet\_b3',
`efficientnet\_b4',
`tf\_efficientnet\_b2',
`tf\_efficientnet\_lite1',
`convnext\_base',
`convnext\_small',
`resnetrs350',  
`pit\_xs\_distilled\_224', 
`crossvit\_small\_240', 
`botnet26t\_256',
`tinynet\_e',
`tinynet\_d',
`repvgg\_b2g4',
`mnasnet\_small',
`dla46x\_c',
`lcnet\_050',
`tv\_resnet34',
`tv\_resnet50',
`tv\_resnet101',
`tv\_resnet152',
`densenet121',
`inception\_v4',
`resnet26d',
`mobilenetv2\_140',
`hrnet\_w40',
`xception',
`xception41',
`resnet18',
`resnet34',
`seresnet50',
`mobilenetv2\_050',
`seresnet33ts',
`wide\_resnet50\_2',
`wide\_resnet101\_2',
`resnet18d',
`hrnet\_w18\_small',
`gluon\_resnet152\_v1d',
`hrnet\_w48',
`hrnet\_w44',
`repvgg\_b2',
`densenet201',
`hrnet\_w18\_small',
`resnet101d',
`gluon\_resnet101\_v1d',
`gluon\_resnet101\_v1s',
`gluon\_xception65',
`gluon\_seresnext50\_32x4d',
`gluon\_senet154',
`gluon\_inception\_v3',
`gluon\_resnet101\_v1c',
`tf\_inception\_v3',
`tv\_densenet121',
`tv\_resnext50\_32x4d',
`repvgg\_b1g4',
`resnext26ts',
`ghostnet\_100',
`crossvit\_9\_240',
`deit\_base\_patch16\_384',
`rexnet\_150',
`rexnet\_130',
`resnetrs50',
`resnet50d',
`resnet50',
`resnetv2\_50',
`resnetrs152',
`resnetrs101',
`dpn92',
`dpn98',
`dpn68',
`vgg19\_bn',
`vgg16\_bn',
`vgg13\_bn',
`vgg11\_bn',
`vgg11',
`vgg11\_bn',
`vgg16',
`vgg19',
`swin\_small\_patch4\_window7\_224',
`swin\_base\_patch4\_window12\_384',
`deit\_base\_patch16\_224',
`deit\_small\_distilled\_patch16\_224',
`densenet161',
`tf\_mobilenetv3\_large\_075',
`inception\_v3'\}} 

\textbf{(2) Semi-supervised learning}
\\
\emph{\{`ssl\_resnext101\_32x8d',
`ssl\_resnext101\_32x16d',
`swsl\_resnext101\_32x8d',
`swsl\_resnext101\_32x16d',
`ssl\_resnext101\_32x4d',
`ssl\_resnext50\_32x4d',
`ssl\_resnet50',
`swsl\_resnext101\_32x4d',
`swsl\_resnext50\_32x4d',
`swsl\_resnet50',
`tf\_efficientnet\_l2\_ns\_475',
`tf\_efficientnet\_b7\_ns',
`tf\_efficientnet\_b6\_ns,
`tf\_efficientnet\_b4\_ns,
`tf\_efficientnet\_b5\_ns'\}} 

\textbf{(3) Pretraining on more data} 
\\
\emph{\{`convnext\_xlarge\_384\_in22ft1k',
`convnext\_xlarge\_in22ft1k', 
`convnext\_large\_384\_in22ft1k',
`convnext\_large\_in22ft1k',
`convnext\_base\_384\_in22ft1k', 
`convnext\_base\_in22ft1k', 
`resnetv2\_152x2\_bitm', 
`resnetv2\_152x4\_bitm',
`resnetv2\_50x1\_bitm',
`resmlp\_big\_24\_224\_in22ft1k',
`resmlp\_big\_24\_distilled\_224', 
`tf\_efficientnetv2\_s\_in21ft1k',
`tf\_efficientnetv2\_m\_in21ft1k',
`tf\_efficientnetv2\_l\_in21ft1k', 
`tf\_efficientnetv2\_xl\_in21ft1k',
`vit\_large\_patch16\_384',
`swin\_large\_patch4\_window12\_384',
`beit\_large\_patch16\_512',
`beit\_large\_patch16\_384',
`beit\_large\_patch16\_224',
`beit\_base\_patch16\_384', 
`vit\_base\_patch16\_384', 
`vit\_small\_r26\_s32\_384',
`vit\_tiny\_patch16\_384',
`vit\_large\_r50\_s32\_384',
`mixer\_b16\_224\_miil',
`resmlp\_big\_24\_224',
`resnetv2\_50x1\_bit\_distilled',
`ig\_resnext101\_32x16d',
`ig\_resnext101\_32x32d',
`ig\_resnext101\_32x8d',
`ig\_resnext101\_32x48d',
`resnext101\_32x16d\_wsl',
`regnety\_16gf\_in1k',
`regnety\_32gf\_in1k’, %
\}}

\textbf{(4) ResNet-50 models trained with diverse augmentation}

(A) Models trained with augmentation policy:\\
`resnet50\_randaa': \textcolor{blue}{https://github.com/zhanghang1989/Fast-AutoAug-Torch};
\\
`resnet50\_aa': \textcolor{blue}{https://github.com/zhanghang1989/Fast-AutoAug-Torch};
\\
`resnet50\_fastaa': \textcolor{blue}{https://github.com/zhanghang1989/Fast-AutoAug-Torch};

(B) Models trained with robust-related data augmentation:\\
`resnet50\_augmix': \textcolor{blue}{https://github.com/google-research/augmix};\\
`resnet50\_cutmix': \textcolor{blue}{https://github.com/clovaai/CutMix-PyTorch}'
\\
`resnet50\_feature\_cutmix': \textcolor{blue}{https://github.com/clovaai/CutMix-PyTorch}'
\\
`resnet50\_deepaugment': \textcolor{blue}{https://github.com/hendrycks/imagenet-r};
\\
`resnet50\_deepaugment\_and\_augmix': \textcolor{blue}{https://github.com/hendrycks/imagenet-r};
\\
`resnet50\_pixmix': \textcolor{blue}{https://github.com/andyzoujm/pixmix};

(C) Adversarially ``robust" models:\\
`resnet50\_l2\_eps1': \textcolor{blue}{https://github.com/microsoft/robust-models-transfer};\\
`resnet50\_l2\_eps0\_5': \textcolor{blue}{https://github.com/microsoft/robust-models-transfer};\\
`resnet50\_l2\_eps0\_25': \textcolor{blue}{https://github.com/microsoft/robust-models-transfer};\\
`resnet50\_l2\_eps0\_03': \textcolor{blue}{https://github.com/microsoft/robust-models-transfer};\\
`resnet50\_l2\_eps0\_01': \textcolor{blue}{https://github.com/microsoft/robust-models-transfer}.

\subsection{CIFAR-10 Models}
Follow the practice in \cite{miller2021accuracy}, we train CIFAR models using the implementations from \textcolor{blue}{https: //github.com/kuangliu/pytorch-cifar}. 
The models span a range of manually designed architectures and the results of automated architecture searches. Specifically, we use \{\emph{`DenseNet-121/169/201/161/201', 
`Densenet-cifar',
`DLA',
`DPN26/92',
`EfficientNetB0',
`GoogLeNet',
`LeNet',
`MobileNet',
`MobileNetV2',
`PNASNetA/B',
`PreActResNet18/34/50/101/152', 
`RegNetX-200MF/400MF',
`RegNetY-400MF',
`ResNet-18/34/50/101/152', 
`ResNeXt29-8x64d/32x4d/4x64d/2x64d',
`SENet18',
`ShuffleNetV2',
`ShuffleNetG2/G3', 
`SimpleDLA',
`VGG-11/13/16/19'}\}.

Furthermore, we use the trained models publicly provided by \textcolor{blue}{https://github.com/chenyaofo/pytorch-cifar-models}. There are \emph{\{`cifar10-mobilenetv2-x0-5',
`cifar10-mobilenetv2-x0-75',
`cifar10-mobilenetv2-x1-0',
`cifar10-mobilenetv2-x1-4',
`cifar10-repvgg-a0',
`cifar10-repvgg-a1',
`cifar10-repvgg-a2',
`cifar10-resnet20',
`cifar10-resnet32',
`cifar10-resnet44',
`cifar10-resnet56',
`cifar10-shufflenetv2-x0-5',
`cifar10-shufflenetv2-x1-0',
`cifar10-shufflenetv2-x1-5',
`cifar10-shufflenetv2-x2-0',
`cifar10-vgg11-bn',
`cifar10-vgg13-bn',
`cifar10-vgg16-bn',
`cifar10-vgg19-bn'\}
}.

\subsection{Datasets} 
The datasets we use are standard benchmarks, which are publicly available. We have double-checked their license. We list their open-source as follows.

\textbf{CIFAR-10} \cite{krizhevsky2009learning} (\textcolor{blue}{https://www.cs.toronto.edu/~kriz/cifar.html});\\
\textbf{CIFAR-10-C} \cite{hendrycks2019robustness} (\textcolor{blue}{https://github.com/hendrycks/robustness}); \\
\textbf{CIFAR-10.1} \cite{recht2018cifar} (\textcolor{blue}{https://github.com/modestyachts/CIFAR-10.1});\\
\textbf{CINIC} \cite{chrabaszcz2017downsampled} (\textcolor{blue}{https://github.com/BayesWatch/cinic-10}).

\textbf{ImageNet-Validation} \cite{deng2009imagenet} (\textcolor{blue}{https://www.image-net.org}); \\
\textbf{ImageNet-V2-A/B/C} \cite{recht2019imagenet} (\textcolor{blue}{https://github.com/modestyachts/ImageNetV2}); \\
\textbf{ImageNet-Corruption} \cite{hendrycks2019robustness} (\textcolor{blue}{https://github.com/hendrycks/robustness});\\
\textbf{ImageNet-Sketch} \cite{wang2019learning} (\textcolor{blue}{https://github.com/HaohanWang/ImageNet-Sketch});\\
\textbf{ImageNet-Adversarial} \cite{hendrycks2021natural} (\textcolor{blue}{https://github.com/hendrycks/natural-adv-examples});\\
\textbf{ImageNet-Rendition} \cite{hendrycks2021many} (\textcolor{blue}{https://github.com/hendrycks/imagenet-r}).

\subsection{Computation Resources} 
We run all experiment on one 3090Ti. CPU is \textcolor{black}{AMD Ryzen 9 5900X 12-Core Processor}. Moreover, PyTorch version is \textcolor{black}{1.11.0+cu113} and timm version is \textcolor{black}{1.5}.


\newpage
\begin{small}
\bibliographystyle{unsrtnat}
\bibliography{egbib}

\begin{thebibliography}{96}
\providecommand{\natexlab}[1]{#1}
\providecommand{\url}[1]{\texttt{#1}}
\expandafter\ifx\csname urlstyle\endcsname\relax
  \providecommand{\doi}[1]{doi: #1}\else
  \providecommand{\doi}{doi: \begingroup \urlstyle{rm}\Url}\fi

\bibitem[Ben-David et~al.(2010)Ben-David, Blitzer, Crammer, Kulesza, Pereira,
  and Vaughan]{shai2010hdh}
Shai Ben-David, John Blitzer, Koby Crammer, Alex Kulesza, Fernando Pereira, and
  Jennifer Vaughan.
\newblock A theory of learning from different domains.
\newblock \emph{Machine Learning}, 79:\penalty0 151--175, 2010.

\bibitem[Miller et~al.(2021)Miller, Taori, Raghunathan, Sagawa, Koh, Shankar,
  Liang, Carmon, and Schmidt]{miller2021accuracy}
John~P Miller, Rohan Taori, Aditi Raghunathan, Shiori Sagawa, Pang~Wei Koh,
  Vaishaal Shankar, Percy Liang, Yair Carmon, and Ludwig Schmidt.
\newblock Accuracy on the line: on the strong correlation between
  out-of-distribution and in-distribution generalization.
\newblock In \emph{International Conference on Machine Learning}, pages
  7721--7735, 2021.

\bibitem[Taori et~al.(2020)Taori, Dave, Shankar, Carlini, Recht, and
  Schmidt]{taori2020measuring}
Rohan Taori, Achal Dave, Vaishaal Shankar, Nicholas Carlini, Benjamin Recht,
  and Ludwig Schmidt.
\newblock Measuring robustness to natural distribution shifts in image
  classification.
\newblock \emph{Advances in Neural Information Processing Systems},
  33:\penalty0 18583--18599, 2020.

\bibitem[Hendrycks et~al.(2021{\natexlab{a}})Hendrycks, Basart, Mu, Kadavath,
  Wang, Dorundo, Desai, Zhu, Parajuli, Guo, et~al.]{hendrycks2021many}
Dan Hendrycks, Steven Basart, Norman Mu, Saurav Kadavath, Frank Wang, Evan
  Dorundo, Rahul Desai, Tyler Zhu, Samyak Parajuli, Mike Guo, et~al.
\newblock The many faces of robustness: A critical analysis of
  out-of-distribution generalization.
\newblock In \emph{Proceedings of the IEEE/CVF International Conference on
  Computer Vision}, pages 8340--8349, 2021{\natexlab{a}}.

\bibitem[Zhang et~al.(2021)Zhang, Bengio, Hardt, Recht, and
  Vinyals]{zhang2021understanding}
Chiyuan Zhang, Samy Bengio, Moritz Hardt, Benjamin Recht, and Oriol Vinyals.
\newblock Understanding deep learning (still) requires rethinking
  generalization.
\newblock \emph{Communications of the ACM}, 64\penalty0 (3):\penalty0 107--115,
  2021.

\bibitem[Azulay and Weiss(2018)]{azulay2018deep}
Aharon Azulay and Yair Weiss.
\newblock Why do deep convolutional networks generalize so poorly to small
  image transformations?
\newblock \emph{arXiv preprint arXiv:1805.12177}, 2018.

\bibitem[Engstrom et~al.(2019)Engstrom, Tran, Tsipras, Schmidt, and
  Madry]{engstrom2019exploring}
Logan Engstrom, Brandon Tran, Dimitris Tsipras, Ludwig Schmidt, and Aleksander
  Madry.
\newblock Exploring the landscape of spatial robustness.
\newblock In \emph{International Conference on Machine Learning}, pages
  1802--1811, 2019.

\bibitem[Kanbak et~al.(2018)Kanbak, Moosavi-Dezfooli, and
  Frossard]{kanbak2018geometric}
Can Kanbak, Seyed-Mohsen Moosavi-Dezfooli, and Pascal Frossard.
\newblock Geometric robustness of deep networks: analysis and improvement.
\newblock In \emph{Proceedings of the IEEE Conference on Computer Vision and
  Pattern Recognition}, pages 4441--4449, 2018.

\bibitem[Zhang(2019)]{zhang2019making}
Richard Zhang.
\newblock Making convolutional networks shift-invariant again.
\newblock In \emph{International conference on machine learning}, pages
  7324--7334, 2019.

\bibitem[Kayhan and Gemert(2020)]{kayhan2020translation}
Osman~Semih Kayhan and Jan C~van Gemert.
\newblock On translation invariance in cnns: Convolutional layers can exploit
  absolute spatial location.
\newblock In \emph{Proceedings of the IEEE/CVF Conference on Computer Vision
  and Pattern Recognition}, pages 14274--14285, 2020.

\bibitem[Zhu et~al.(2021)Zhu, An, and Huang]{zhu2021understanding}
Sicheng Zhu, Bang An, and Furong Huang.
\newblock Understanding the generalization benefit of model invariance from a
  data perspective.
\newblock In \emph{Advances in Neural Information Processing Systems},
  volume~34, pages 4328--4341, 2021.

\bibitem[Zhou et~al.(2017)Zhou, Ye, Qiu, and Jiao]{zhou2017oriented}
Yanzhao Zhou, Qixiang Ye, Qiang Qiu, and Jianbin Jiao.
\newblock Oriented response networks.
\newblock In \emph{Proceedings of the IEEE Conference on Computer Vision and
  Pattern Recognition}, pages 519--528, 2017.

\bibitem[Jaderberg et~al.(2015)Jaderberg, Simonyan, Zisserman,
  et~al.]{jaderberg2015spatial}
Max Jaderberg, Karen Simonyan, Andrew Zisserman, et~al.
\newblock Spatial transformer networks.
\newblock In \emph{Advances in neural information processing systems}, 2015.

\bibitem[Delchevalerie et~al.(2021)Delchevalerie, Bibal, Fr{\'e}nay, and
  Mayer]{delchevalerie2021achieving}
Valentin Delchevalerie, Adrien Bibal, Beno{\^\i}t Fr{\'e}nay, and Alexandre
  Mayer.
\newblock Achieving rotational invariance with bessel-convolutional neural
  networks.
\newblock In \emph{Advances in Neural Information Processing Systems}, 2021.

\bibitem[Recht et~al.(2019)Recht, Roelofs, Schmidt, and
  Shankar]{recht2019imagenet}
Benjamin Recht, Rebecca Roelofs, Ludwig Schmidt, and Vaishaal Shankar.
\newblock Do imagenet classifiers generalize to imagenet?
\newblock In \emph{International Conference on Machine Learning}, pages
  5389--5400. PMLR, 2019.

\bibitem[Schiff et~al.(2021)Schiff, Quanz, Das, and Chen]{schiff2021predicting}
Yair Schiff, Brian Quanz, Payel Das, and Pin-Yu Chen.
\newblock Predicting deep neural network generalization with perturbation
  response curves.
\newblock In \emph{Advances in Neural Information Processing Systems}, 2021.

\bibitem[Deng et~al.(2009)Deng, Dong, Socher, Li, Li, and
  Fei-Fei]{deng2009imagenet}
Jia Deng, Wei Dong, Richard Socher, Li-Jia Li, Kai Li, and Li~Fei-Fei.
\newblock Imagenet: A large-scale hierarchical image database.
\newblock In \emph{Proceedings of the IEEE Conference on Computer Vision and
  Pattern Recognition}, pages 248--255, 2009.

\bibitem[Simonyan and Zisserman(2014)]{simonyan2014very}
Karen Simonyan and Andrew Zisserman.
\newblock Very deep convolutional networks for large-scale image recognition.
\newblock \emph{arXiv preprint arXiv:1409.1556}, 2014.

\bibitem[Bao et~al.(2021)Bao, Dong, and Wei]{bao2021beit}
Hangbo Bao, Li~Dong, and Furu Wei.
\newblock Beit: Bert pre-training of image transformers.
\newblock \emph{arXiv preprint arXiv:2106.08254}, 2021.

\bibitem[Eilertsen et~al.(2020)Eilertsen, J{\"o}nsson, Ropinski, Unger, and
  Ynnerman]{eilertsen2020classifying}
Gabriel Eilertsen, Daniel J{\"o}nsson, Timo Ropinski, Jonas Unger, and Anders
  Ynnerman.
\newblock Classifying the classifier: dissecting the weight space of neural
  networks.
\newblock \emph{arXiv preprint arXiv:2002.05688}, 2020.

\bibitem[Unterthiner et~al.(2020)Unterthiner, Keysers, Gelly, Bousquet, and
  Tolstikhin]{unterthiner2020predicting}
Thomas Unterthiner, Daniel Keysers, Sylvain Gelly, Olivier Bousquet, and Ilya
  Tolstikhin.
\newblock Predicting neural network accuracy from weights.
\newblock In \emph{International Conference on Learning Representations}, 2020.

\bibitem[Arora et~al.(2018)Arora, Ge, Neyshabur, and Zhang]{arora2018stronger}
Sanjeev Arora, Rong Ge, Behnam Neyshabur, and Yi~Zhang.
\newblock Stronger generalization bounds for deep nets via a compression
  approach.
\newblock \emph{arXiv preprint arXiv:1802.05296}, 2018.

\bibitem[Corneanu et~al.(2020)Corneanu, Escalera, and
  Martinez]{corneanu2020computing}
Ciprian~A Corneanu, Sergio Escalera, and Aleix~M Martinez.
\newblock Computing the testing error without a testing set.
\newblock In \emph{Proceedings of the IEEE conference on computer vision and
  pattern recognition}, pages 2677--2685, 2020.

\bibitem[Jiang et~al.(2019{\natexlab{a}})Jiang, Krishnan, Mobahi, and
  Bengio]{jiang2018predicting}
Yiding Jiang, Dilip Krishnan, Hossein Mobahi, and Samy Bengio.
\newblock Predicting the generalization gap in deep networks with margin
  distributions.
\newblock In \emph{International Conference on Learning Representations},
  2019{\natexlab{a}}.

\bibitem[Neyshabur et~al.(2017)Neyshabur, Bhojanapalli, McAllester, and
  Srebro]{neyshabur2017exploring}
Behnam Neyshabur, Srinadh Bhojanapalli, David McAllester, and Nati Srebro.
\newblock Exploring generalization in deep learning.
\newblock In \emph{Advances in neural information processing systems}, pages
  5947--5956, 2017.

\bibitem[Jiang et~al.(2019{\natexlab{b}})Jiang, Neyshabur, Mobahi, Krishnan,
  and Bengio]{jiang2019fantastic}
Yiding Jiang, Behnam Neyshabur, Hossein Mobahi, Dilip Krishnan, and Samy
  Bengio.
\newblock Fantastic generalization measures and where to find them.
\newblock In \emph{International Conference on Learning Representations},
  2019{\natexlab{b}}.

\bibitem[Garg et~al.(2021)Garg, Balakrishnan, Kolter, and Lipton]{garg2021ratt}
Saurabh Garg, Sivaraman Balakrishnan, Zico Kolter, and Zachary Lipton.
\newblock Ratt: Leveraging unlabeled data to guarantee generalization.
\newblock In \emph{International Conference on Machine Learning}, pages
  3598--3609, 2021.

\bibitem[Jiang et~al.(2021)Jiang, Nagarajan, Baek, and
  Kolter]{jiang2021assessing}
Yiding Jiang, Vaishnavh Nagarajan, Christina Baek, and J~Zico Kolter.
\newblock Assessing generalization of sgd via disagreement.
\newblock \emph{arXiv preprint arXiv:2106.13799}, 2021.

\bibitem[Aithal et~al.(2021)Aithal, Kashyap, and
  Subramanyam]{kashyap2021robustness}
Sumukh Aithal, Dhruva Kashyap, and Natarajan Subramanyam.
\newblock Robustness to augmentations as a generalization metric.
\newblock \emph{arXiv preprint arXiv:2101.06459}, 2021.

\bibitem[Deng and Zheng(2021)]{deng2021labels}
Weijian Deng and Liang Zheng.
\newblock Are labels always necessary for classifier accuracy evaluation?
\newblock In \emph{Proceedings of the IEEE/CVF Conference on Computer Vision
  and Pattern Recognition}, pages 15069--15078, 2021.

\bibitem[Guillory et~al.(2021)Guillory, Shankar, Ebrahimi, Darrell, and
  Schmidt]{guillory2021predicting}
Devin Guillory, Vaishaal Shankar, Sayna Ebrahimi, Trevor Darrell, and Ludwig
  Schmidt.
\newblock Predicting with confidence on unseen distributions.
\newblock In \emph{Proceedings of the IEEE/CVF International Conference on
  Computer Vision}, pages 1134--1144, 2021.

\bibitem[Garg et~al.(2022)Garg, Balakrishnan, Lipton, Neyshabur, and
  Sedghi]{garg2022leveraging}
Saurabh Garg, Sivaraman Balakrishnan, Zachary~C Lipton, Behnam Neyshabur, and
  Hanie Sedghi.
\newblock Leveraging unlabeled data to predict out-of-distribution performance.
\newblock In \emph{International Conference on Learning Representations}, 2022.

\bibitem[Baek et~al.(2022)Baek, Jiang, Raghunathan, and
  Kolter]{baek2022agreement}
Christina Baek, Yiding Jiang, Aditi Raghunathan, and Zico Kolter.
\newblock Agreement-on-the-line: Predicting the performance of neural networks
  under distribution shift.
\newblock \emph{arXiv preprint arXiv:2206.13089}, 2022.

\bibitem[Ng et~al.(2022)Ng, Cho, Hulkund, and Ghassemi]{ng2022predicting}
Nathan Ng, Kyunghyun Cho, Neha Hulkund, and Marzyeh Ghassemi.
\newblock Predicting out-of-domain generalization with local manifold
  smoothness.
\newblock \emph{arXiv preprint arXiv:2207.02093}, 2022.

\bibitem[Yu et~al.(2022)Yu, Yang, Wei, Ma, and Steinhardt]{yu2022predicting}
Yaodong Yu, Zitong Yang, Alexander Wei, Yi~Ma, and Jacob Steinhardt.
\newblock Predicting out-of-distribution error with the projection norm.
\newblock In \emph{Advances in Neural Information Processing Systems}, 2022.

\bibitem[Deng et~al.(2021)Deng, Gould, and Zheng]{Deng:ICML2021}
Weijian Deng, Stephen Gould, and Liang Zheng.
\newblock What does rotation prediction tell us about classifier accuracy under
  varying testing environments?
\newblock In \emph{International conference on machine learning}, 2021.

\bibitem[Chen et~al.(2021)Chen, Liu, Avci, Wu, Liang, and
  Jha]{chen2021detecting}
Jiefeng Chen, Frederick Liu, Besim Avci, Xi~Wu, Yingyu Liang, and Somesh Jha.
\newblock Detecting errors and estimating accuracy on unlabeled data with
  self-training ensembles.
\newblock \emph{Advances in Neural Information Processing Systems}, 34, 2021.

\bibitem[Hendrycks et~al.(2019)Hendrycks, Mu, Cubuk, Zoph, Gilmer, and
  Lakshminarayanan]{hendrycks2019augmix}
Dan Hendrycks, Norman Mu, Ekin~D Cubuk, Barret Zoph, Justin Gilmer, and Balaji
  Lakshminarayanan.
\newblock Augmix: A simple data processing method to improve robustness and
  uncertainty.
\newblock \emph{arXiv preprint arXiv:1912.02781}, 2019.

\bibitem[Yun et~al.(2019)Yun, Han, Oh, Chun, Choe, and Yoo]{yun2019cutmix}
Sangdoo Yun, Dongyoon Han, Seong~Joon Oh, Sanghyuk Chun, Junsuk Choe, and
  Youngjoon Yoo.
\newblock Cutmix: Regularization strategy to train strong classifiers with
  localizable features.
\newblock In \emph{Proceedings of the IEEE/CVF international conference on
  computer vision}, pages 6023--6032, 2019.

\bibitem[Cubuk et~al.(2018)Cubuk, Zoph, Mane, Vasudevan, and
  Le]{cubuk2018autoaugment}
Ekin~D Cubuk, Barret Zoph, Dandelion Mane, Vijay Vasudevan, and Quoc~V Le.
\newblock Autoaugment: Learning augmentation policies from data.
\newblock \emph{arXiv preprint arXiv:1805.09501}, 2018.

\bibitem[Cubuk et~al.(2020)Cubuk, Zoph, Shlens, and Le]{cubuk2020randaugment}
Ekin~D Cubuk, Barret Zoph, Jonathon Shlens, and Quoc~V Le.
\newblock Randaugment: Practical automated data augmentation with a reduced
  search space.
\newblock In \emph{Proceedings of the IEEE/CVF Conference on Computer Vision
  and Pattern Recognition Workshops}, pages 702--703, 2020.

\bibitem[DeVries and Taylor(2017)]{devries2017improved}
Terrance DeVries and Graham~W Taylor.
\newblock Improved regularization of convolutional neural networks with cutout.
\newblock \emph{arXiv preprint arXiv:1708.04552}, 2017.

\bibitem[Mintun et~al.(2021)Mintun, Kirillov, and Xie]{mintun2021interaction}
Eric Mintun, Alexander Kirillov, and Saining Xie.
\newblock On interaction between augmentations and corruptions in natural
  corruption robustness.
\newblock In \emph{Advances in Neural Information Processing Systems}, 2021.

\bibitem[Hendrycks et~al.(2022)Hendrycks, Zou, Mazeika, Tang, Song, and
  Steinhardt]{hendrycks2021pixmix}
Dan Hendrycks, Andy Zou, Mantas Mazeika, Leonard Tang, Dawn Song, and Jacob
  Steinhardt.
\newblock Pixmix: Dreamlike pictures comprehensively improve safety measures.
\newblock In \emph{Proceedings of the IEEE Conference on Computer Vision and
  Pattern Recognition}, 2022.

\bibitem[Zhang et~al.(2017)Zhang, Cisse, Dauphin, and
  Lopez-Paz]{zhang2017mixup}
Hongyi Zhang, Moustapha Cisse, Yann~N Dauphin, and David Lopez-Paz.
\newblock mixup: Beyond empirical risk minimization.
\newblock \emph{arXiv preprint arXiv:1710.09412}, 2017.

\bibitem[Tokozume et~al.(2018)Tokozume, Ushiku, and
  Harada]{tokozume2018between}
Yuji Tokozume, Yoshitaka Ushiku, and Tatsuya Harada.
\newblock Between-class learning for image classification.
\newblock In \emph{Proceedings of the IEEE Conference on Computer Vision and
  Pattern Recognition}, pages 5486--5494, 2018.

\bibitem[Yin et~al.(2019)Yin, Gontijo~Lopes, Shlens, Cubuk, and
  Gilmer]{yin2019fourier}
Dong Yin, Raphael Gontijo~Lopes, Jon Shlens, Ekin~Dogus Cubuk, and Justin
  Gilmer.
\newblock A fourier perspective on model robustness in computer vision.
\newblock In \emph{Advances in Neural Information Processing Systems}, 2019.

\bibitem[Madry et~al.(2017)Madry, Makelov, Schmidt, Tsipras, and
  Vladu]{madry2017towards}
Aleksander Madry, Aleksandar Makelov, Ludwig Schmidt, Dimitris Tsipras, and
  Adrian Vladu.
\newblock Towards deep learning models resistant to adversarial attacks.
\newblock \emph{arXiv preprint arXiv:1706.06083}, 2017.

\bibitem[Salman et~al.(2020)Salman, Ilyas, Engstrom, Kapoor, and
  Madry]{salman2020adversarially}
Hadi Salman, Andrew Ilyas, Logan Engstrom, Ashish Kapoor, and Aleksander Madry.
\newblock Do adversarially robust imagenet models transfer better?
\newblock In \emph{Advances in Neural Information Processing Systems}, pages
  3533--3545, 2020.

\bibitem[Rusak et~al.(2020)Rusak, Schott, Zimmermann, Bitterwolf, Bringmann,
  Bethge, and Brendel]{rusak2020simple}
Evgenia Rusak, Lukas Schott, Roland~S Zimmermann, Julian Bitterwolf, Oliver
  Bringmann, Matthias Bethge, and Wieland Brendel.
\newblock A simple way to make neural networks robust against diverse image
  corruptions.
\newblock In \emph{European Conference on Computer Vision}, pages 53--69, 2020.

\bibitem[Fuglede and Topsoe(2004)]{fuglede2004jensen}
Bent Fuglede and Flemming Topsoe.
\newblock Jensen-shannon divergence and hilbert space embedding.
\newblock In \emph{International Symposium onInformation Theory, 2004. ISIT
  2004. Proceedings.}, page~31, 2004.

\bibitem[Sohn et~al.(2020)Sohn, Berthelot, Carlini, Zhang, Zhang, Raffel,
  Cubuk, Kurakin, and Li]{sohn2020fixmatch}
Kihyuk Sohn, David Berthelot, Nicholas Carlini, Zizhao Zhang, Han Zhang,
  Colin~A Raffel, Ekin~Dogus Cubuk, Alexey Kurakin, and Chun-Liang Li.
\newblock Fixmatch: Simplifying semi-supervised learning with consistency and
  confidence.
\newblock In \emph{Advances in Neural Information Processing Systems}, pages
  596--608, 2020.

\bibitem[Berthelot et~al.(2019)Berthelot, Carlini, Goodfellow, Papernot,
  Oliver, and Raffel]{berthelot2019mixmatch}
David Berthelot, Nicholas Carlini, Ian Goodfellow, Nicolas Papernot, Avital
  Oliver, and Colin~A Raffel.
\newblock Mixmatch: A holistic approach to semi-supervised learning.
\newblock In \emph{Advances in Neural Information Processing Systems}, 2019.

\bibitem[Kullback and Leibler(1951)]{kullback1951information}
Solomon Kullback and Richard~A Leibler.
\newblock On information and sufficiency.
\newblock \emph{The annals of mathematical statistics}, 22\penalty0
  (1):\penalty0 79--86, 1951.

\bibitem[Sun et~al.(2021)Sun, Mehra, Kailkhura, Chen, Hendrycks, Hamm, and
  Mao]{sun2021certified}
Jiachen Sun, Akshay Mehra, Bhavya Kailkhura, Pin-Yu Chen, Dan Hendrycks, Jihun
  Hamm, and Z~Morley Mao.
\newblock Certified adversarial defenses meet out-of-distribution corruptions:
  Benchmarking robustness and simple baselines.
\newblock \emph{arXiv preprint arXiv:2112.00659}, 2021.

\bibitem[He et~al.(2016)He, Zhang, Ren, and Sun]{he2016deep}
Kaiming He, Xiangyu Zhang, Shaoqing Ren, and Jian Sun.
\newblock Deep residual learning for image recognition.
\newblock In \emph{Proceedings of the IEEE conference on computer vision and
  pattern recognition}, pages 770--778, 2016.

\bibitem[Liu et~al.(2022)Liu, Mao, Wu, Feichtenhofer, Darrell, and
  Xie]{liu2022convnet}
Zhuang Liu, Hanzi Mao, Chao-Yuan Wu, Christoph Feichtenhofer, Trevor Darrell,
  and Saining Xie.
\newblock A convnet for the 2020s.
\newblock In \emph{Proceedings of the IEEE Conference on Computer Vision and
  Pattern Recognition}, 2022.

\bibitem[Dosovitskiy et~al.(2020)Dosovitskiy, Beyer, Kolesnikov, Weissenborn,
  Zhai, Unterthiner, Dehghani, Minderer, Heigold, Gelly,
  et~al.]{dosovitskiy2020image}
Alexey Dosovitskiy, Lucas Beyer, Alexander Kolesnikov, Dirk Weissenborn,
  Xiaohua Zhai, Thomas Unterthiner, Mostafa Dehghani, Matthias Minderer, Georg
  Heigold, Sylvain Gelly, et~al.
\newblock An image is worth 16x16 words: Transformers for image recognition at
  scale.
\newblock \emph{arXiv preprint arXiv:2010.11929}, 2020.

\bibitem[Liu et~al.(2021)Liu, Lin, Cao, Hu, Wei, Zhang, Lin, and
  Guo]{liu2021swin}
Ze~Liu, Yutong Lin, Yue Cao, Han Hu, Yixuan Wei, Zheng Zhang, Stephen Lin, and
  Baining Guo.
\newblock Swin transformer: Hierarchical vision transformer using shifted
  windows.
\newblock In \emph{Proceedings of the IEEE/CVF International Conference on
  Computer Vision}, pages 10012--10022, 2021.

\bibitem[Ding et~al.(2022)Ding, Xia, Zhang, Chu, Han, and Ding]{ding2021repmlp}
Xiaohan Ding, Chunlong Xia, Xiangyu Zhang, Xiaojie Chu, Jungong Han, and
  Guiguang Ding.
\newblock Repmlp: Re-parameterizing convolutions into fully-connected layers
  for image recognition.
\newblock In \emph{Proceedings of the IEEE/CVF conference on computer vision
  and pattern recognition}, 2022.

\bibitem[Tolstikhin et~al.(2021)Tolstikhin, Houlsby, Kolesnikov, Beyer, Zhai,
  Unterthiner, Yung, Steiner, Keysers, Uszkoreit, et~al.]{tolstikhin2021mlp}
Ilya~O Tolstikhin, Neil Houlsby, Alexander Kolesnikov, Lucas Beyer, Xiaohua
  Zhai, Thomas Unterthiner, Jessica Yung, Andreas Steiner, Daniel Keysers,
  Jakob Uszkoreit, et~al.
\newblock Mlp-mixer: An all-mlp architecture for vision.
\newblock In \emph{Advances in Neural Information Processing Systems}, 2021.

\bibitem[Goyal et~al.(2017)Goyal, Doll{\'a}r, Girshick, Noordhuis, Wesolowski,
  Kyrola, Tulloch, Jia, and He]{goyal2017accurate}
Priya Goyal, Piotr Doll{\'a}r, Ross Girshick, Pieter Noordhuis, Lukasz
  Wesolowski, Aapo Kyrola, Andrew Tulloch, Yangqing Jia, and Kaiming He.
\newblock Accurate, large minibatch sgd: Training imagenet in 1 hour.
\newblock \emph{arXiv preprint arXiv:1706.02677}, 2017.

\bibitem[Szegedy et~al.(2016)Szegedy, Vanhoucke, Ioffe, Shlens, and
  Wojna]{szegedy2016rethinking}
Christian Szegedy, Vincent Vanhoucke, Sergey Ioffe, Jon Shlens, and Zbigniew
  Wojna.
\newblock Rethinking the inception architecture for computer vision.
\newblock In \emph{Proceedings of the IEEE conference on computer vision and
  pattern recognition}, pages 2818--2826, 2016.

\bibitem[Bello et~al.(2021)Bello, Fedus, Du, Cubuk, Srinivas, Lin, Shlens, and
  Zoph]{bello2021revisiting}
Irwan Bello, William Fedus, Xianzhi Du, Ekin~Dogus Cubuk, Aravind Srinivas,
  Tsung-Yi Lin, Jonathon Shlens, and Barret Zoph.
\newblock Revisiting resnets: Improved training and scaling strategies.
\newblock \emph{Advances in Neural Information Processing Systems}, 34, 2021.

\bibitem[Tan and Le(2019)]{tan2019efficientnet}
Mingxing Tan and Quoc Le.
\newblock Efficientnet: Rethinking model scaling for convolutional neural
  networks.
\newblock In \emph{International conference on machine learning}, pages
  6105--6114, 2019.

\bibitem[Tan and Le(2021)]{tan2021efficientnetv2}
Mingxing Tan and Quoc Le.
\newblock Efficientnetv2: Smaller models and faster training.
\newblock In \emph{International Conference on Machine Learning}, pages
  10096--10106, 2021.

\bibitem[Yalniz et~al.(2019)Yalniz, J{\'e}gou, Chen, Paluri, and
  Mahajan]{yalniz2019billion}
I~Zeki Yalniz, Herv{\'e} J{\'e}gou, Kan Chen, Manohar Paluri, and Dhruv
  Mahajan.
\newblock Billion-scale semi-supervised learning for image classification.
\newblock \emph{arXiv preprint arXiv:1905.00546}, 2019.

\bibitem[Hinton et~al.(2015)Hinton, Vinyals, Dean,
  et~al.]{hinton2015distilling}
Geoffrey Hinton, Oriol Vinyals, Jeff Dean, et~al.
\newblock Distilling the knowledge in a neural network.
\newblock \emph{arXiv preprint arXiv:1503.02531}, 2\penalty0 (7), 2015.

\bibitem[Heo et~al.(2021)Heo, Yun, Han, Chun, Choe, and Oh]{heo2021rethinking}
Byeongho Heo, Sangdoo Yun, Dongyoon Han, Sanghyuk Chun, Junsuk Choe, and
  Seong~Joon Oh.
\newblock Rethinking spatial dimensions of vision transformers.
\newblock In \emph{Proceedings of the IEEE/CVF International Conference on
  Computer Vision}, pages 11936--11945, 2021.

\bibitem[Wightman(2019)]{rw2019timm}
Ross Wightman.
\newblock Pytorch image models.
\newblock \url{https://github.com/rwightman/pytorch-image-models}, 2019.

\bibitem[Thomee et~al.(2015)Thomee, Shamma, Friedland, Elizalde, Ni, Poland,
  Borth, and Li]{thomee2015new}
Bart Thomee, David~A Shamma, Gerald Friedland, Benjamin Elizalde, Karl Ni,
  Douglas Poland, Damian Borth, and Li-Jia Li.
\newblock The new data and new challenges in multimedia research.
\newblock \emph{arXiv preprint arXiv:1503.01817}, 2015.

\bibitem[Mahajan et~al.(2018)Mahajan, Girshick, Ramanathan, He, Paluri, Li,
  Bharambe, and Van Der~Maaten]{mahajan2018exploring}
Dhruv Mahajan, Ross Girshick, Vignesh Ramanathan, Kaiming He, Manohar Paluri,
  Yixuan Li, Ashwin Bharambe, and Laurens Van Der~Maaten.
\newblock Exploring the limits of weakly supervised pretraining.
\newblock In \emph{Proceedings of the European conference on computer vision
  (ECCV)}, pages 181--196, 2018.

\bibitem[Sun et~al.(2017)Sun, Shrivastava, Singh, and Gupta]{sun2017revisiting}
Chen Sun, Abhinav Shrivastava, Saurabh Singh, and Abhinav Gupta.
\newblock Revisiting unreasonable effectiveness of data in deep learning era.
\newblock In \emph{Proceedings of the IEEE international conference on computer
  vision}, pages 843--852, 2017.

\bibitem[Xie et~al.(2020)Xie, Luong, Hovy, and Le]{xie2020self}
Qizhe Xie, Minh-Thang Luong, Eduard Hovy, and Quoc~V Le.
\newblock Self-training with noisy student improves imagenet classification.
\newblock In \emph{Proceedings of the IEEE/CVF conference on computer vision
  and pattern recognition}, pages 10687--10698, 2020.

\bibitem[Singh et~al.(2022)Singh, Gustafson, Adcock, Reis, Gedik, Kosaraju,
  Mahajan, Girshick, Doll{\'a}r, and van~der Maaten]{singh2022revisiting}
Mannat Singh, Laura Gustafson, Aaron Adcock, Vinicius de~Freitas Reis, Bugra
  Gedik, Raj~Prateek Kosaraju, Dhruv Mahajan, Ross Girshick, Piotr Doll{\'a}r,
  and Laurens van~der Maaten.
\newblock Revisiting weakly supervised pre-training of visual perception
  models.
\newblock In \emph{Proceedings of the IEEE Conference on Computer Vision and
  Pattern Recognition}, 2022.

\bibitem[Kolesnikov et~al.(2020)Kolesnikov, Beyer, Zhai, Puigcerver, Yung,
  Gelly, and Houlsby]{kolesnikov2020big}
Alexander Kolesnikov, Lucas Beyer, Xiaohua Zhai, Joan Puigcerver, Jessica Yung,
  Sylvain Gelly, and Neil Houlsby.
\newblock Big transfer (bit): General visual representation learning.
\newblock In \emph{European conference on computer vision}, pages 491--507,
  2020.

\bibitem[Hendrycks et~al.(2021{\natexlab{b}})Hendrycks, Zhao, Basart,
  Steinhardt, and Song]{hendrycks2021natural}
Dan Hendrycks, Kevin Zhao, Steven Basart, Jacob Steinhardt, and Dawn Song.
\newblock Natural adversarial examples.
\newblock In \emph{Proceedings of the IEEE/CVF Conference on Computer Vision
  and Pattern Recognition}, pages 15262--15271, 2021{\natexlab{b}}.

\bibitem[Wang et~al.(2019)Wang, Ge, Lipton, and Xing]{wang2019learning}
Haohan Wang, Songwei Ge, Zachary Lipton, and Eric~P Xing.
\newblock Learning robust global representations by penalizing local predictive
  power.
\newblock In \emph{Advances in Neural Information Processing Systems}, pages
  10506--10518, 2019.

\bibitem[Hendrycks and Dietterich(2019)]{hendrycks2019robustness}
Dan Hendrycks and Thomas Dietterich.
\newblock Benchmarking neural network robustness to common corruptions and
  perturbations.
\newblock In \emph{Proceedings of the International Conference on Learning
  Representations}, 2019.

\bibitem[Benesty et~al.(2009)Benesty, Chen, Huang, and
  Cohen]{benesty2009pearson}
Jacob Benesty, Jingdong Chen, Yiteng Huang, and Israel Cohen.
\newblock Pearson correlation coefficient.
\newblock In \emph{Noise reduction in speech processing}, pages 1--4. Springer,
  2009.

\bibitem[Kendall(1948)]{kendall1948rank}
Maurice~George Kendall.
\newblock Rank correlation methods.
\newblock 1948.

\bibitem[Huber(2011)]{huber2011robust}
Peter~J Huber.
\newblock Robust statistics.
\newblock In \emph{International encyclopedia of statistical science}, pages
  1248--1251. Springer, 2011.

\bibitem[Meding et~al.(2022)Meding, Buschoff, Geirhos, and
  Wichmann]{meding2021trivial}
Kristof Meding, Luca M~Schulze Buschoff, Robert Geirhos, and Felix~A Wichmann.
\newblock Trivial or impossible--dichotomous data difficulty masks model
  differences (on imagenet and beyond).
\newblock In \emph{Proceedings of the International Conference on Learning
  Representations}, 2022.

\bibitem[Hacohen et~al.(2020)Hacohen, Choshen, and Weinshall]{hacohen2020let}
Guy Hacohen, Leshem Choshen, and Daphna Weinshall.
\newblock Let’s agree to agree: Neural networks share classification order on
  real datasets.
\newblock In \emph{International Conference on Machine Learning}, pages
  3950--3960, 2020.

\bibitem[Yang et~al.(2020)Yang, Qinami, Fei-Fei, Deng, and
  Russakovsky]{yang2020towards}
Kaiyu Yang, Klint Qinami, Li~Fei-Fei, Jia Deng, and Olga Russakovsky.
\newblock Towards fairer datasets: Filtering and balancing the distribution of
  the people subtree in the imagenet hierarchy.
\newblock In \emph{Proceedings of the 2020 Conference on Fairness,
  Accountability, and Transparency}, pages 547--558, 2020.

\bibitem[Recht et~al.(2018)Recht, Roelofs, Schmidt, and
  Shankar]{recht2018cifar}
Benjamin Recht, Rebecca Roelofs, Ludwig Schmidt, and Vaishaal Shankar.
\newblock Do cifar-10 classifiers generalize to cifar-10?
\newblock \emph{arXiv preprint arXiv:1806.00451}, 2018.

\bibitem[Torralba et~al.(2008)Torralba, Fergus, and Freeman]{torralba200880}
Antonio Torralba, Rob Fergus, and William~T Freeman.
\newblock 80 million tiny images: A large data set for nonparametric object and
  scene recognition.
\newblock \emph{IEEE transactions on pattern analysis and machine
  intelligence}, 30\penalty0 (11):\penalty0 1958--1970, 2008.

\bibitem[Chrabaszcz et~al.(2017)Chrabaszcz, Loshchilov, and
  Hutter]{chrabaszcz2017downsampled}
Patryk Chrabaszcz, Ilya Loshchilov, and Frank Hutter.
\newblock A downsampled variant of imagenet as an alternative to the cifar
  datasets.
\newblock \emph{arXiv preprint arXiv:1707.08819}, 2017.

\bibitem[Von~K{\"u}gelgen et~al.(2021)Von~K{\"u}gelgen, Sharma, Gresele,
  Brendel, Sch{\"o}lkopf, Besserve, and Locatello]{von2021self}
Julius Von~K{\"u}gelgen, Yash Sharma, Luigi Gresele, Wieland Brendel, Bernhard
  Sch{\"o}lkopf, Michel Besserve, and Francesco Locatello.
\newblock Self-supervised learning with data augmentations provably isolates
  content from style.
\newblock In \emph{Proceedings of the International Conference on Learning
  Representations}, 2021.

\bibitem[Weiler and Cesa(2019)]{weiler2019general}
Maurice Weiler and Gabriele Cesa.
\newblock General e (2)-equivariant steerable cnns.
\newblock In \emph{Advances in Neural Information Processing Systems}, 2019.

\bibitem[Cohen et~al.(2018)Cohen, Geiger, K{\"o}hler, and
  Welling]{cohen2018spherical}
Taco~S Cohen, Mario Geiger, Jonas K{\"o}hler, and Max Welling.
\newblock Spherical cnns.
\newblock \emph{arXiv preprint arXiv:1801.10130}, 2018.

\bibitem[Koh et~al.(2021)Koh, Sagawa, Marklund, Xie, Zhang, Balsubramani, Hu,
  Yasunaga, Phillips, Gao, Lee, David, Stavness, Guo, Earnshaw, Haque, Beery,
  Leskovec, Kundaje, Pierson, Levine, Finn, and Liang]{wilds2021}
Pang~Wei Koh, Shiori Sagawa, Henrik Marklund, Sang~Michael Xie, Marvin Zhang,
  Akshay Balsubramani, Weihua Hu, Michihiro Yasunaga, Richard~Lanas Phillips,
  Irena Gao, Tony Lee, Etienne David, Ian Stavness, Wei Guo, Berton~A.
  Earnshaw, Imran~S. Haque, Sara Beery, Jure Leskovec, Anshul Kundaje, Emma
  Pierson, Sergey Levine, Chelsea Finn, and Percy Liang.
\newblock {WILDS}: A benchmark of in-the-wild distribution shifts.
\newblock In \emph{International Conference on Machine Learning (ICML)}, 2021.

\bibitem[Ren et~al.(2015)Ren, He, Girshick, and Sun]{ren2015faster}
Shaoqing Ren, Kaiming He, Ross Girshick, and Jian Sun.
\newblock Faster r-cnn: Towards real-time object detection with region proposal
  networks.
\newblock In \emph{Advances in neural information processing systems}, 2015.

\bibitem[Long et~al.(2015)Long, Shelhamer, and Darrell]{long2015fully}
Jonathan Long, Evan Shelhamer, and Trevor Darrell.
\newblock Fully convolutional networks for semantic segmentation.
\newblock In \emph{Proceedings of the IEEE conference on computer vision and
  pattern recognition}, pages 3431--3440, 2015.

\bibitem[Wang et~al.(2018)Wang, Girshick, Gupta, and He]{wang2018non}
Xiaolong Wang, Ross Girshick, Abhinav Gupta, and Kaiming He.
\newblock Non-local neural networks.
\newblock In \emph{Proceedings of the IEEE conference on computer vision and
  pattern recognition}, pages 7794--7803, 2018.

\bibitem[Krizhevsky et~al.(2009)Krizhevsky, Hinton,
  et~al.]{krizhevsky2009learning}
Alex Krizhevsky, Geoffrey Hinton, et~al.
\newblock Learning multiple layers of features from tiny images.
\newblock 2009.

\end{thebibliography}
\end{small}
\end{document}